\documentclass{article}

\usepackage[main,final]{neurips_2026}

\usepackage[utf8]{inputenc}
\usepackage[T1]{fontenc}
\usepackage{hyperref}
\usepackage{url}
\usepackage{booktabs}
\usepackage{amsmath,amssymb,amsfonts}
\usepackage{nicefrac}
\usepackage{microtype}
\usepackage{xcolor}
\usepackage{graphicx}
\usepackage[table]{xcolor}
\usepackage[dvipsnames]{xcolor}
\usepackage{multirow}
\usepackage{my_commands}
\usepackage{makecell}
\usepackage{tabularx}
\usepackage{array}
\usepackage{wrapfig}

\definecolor{darkblue}{rgb}{0.0,0.0,0.45}
\definecolor{royalblue}{rgb}{0.2549,0.4118,0.8824}
\hypersetup{
    colorlinks=true,
    citecolor=royalblue,
    linkcolor=royalblue,
    urlcolor=royalblue
}

\title{PriFT: Guiding Supervised Fine-Tuning with Pretrained Priors}

\author{%
  Ke Wang\thanks{Equal contribution.} \\
    EPFL, Lausanne, Switzerland\\
  \texttt{k.wang@epfl.ch} \\
  \And
  Shuangqi Li\footnotemark[1] \\
    EPFL, Lausanne, Switzerland\\
  \texttt{shuangqi.li@epfl.ch} \\
  \And
  Mathieu Salzmann \\
    EPFL, Lausanne, Switzerland\\
  \texttt{mathieu.salzmann@epfl.ch} \\
  \And
  Pascal Frossard\\
    EPFL, Lausanne, Switzerland\\
  \texttt{pascal.frossard@epfl.ch} 
}

\begin{document}

\maketitle

\begin{abstract}
Supervised fine-tuning (SFT) is an efficient approach for downstream task adaptation and often serves as the initialization stage for reinforcement learning (RL), but it can show weaker generalization than RL. 
A key limitation is its off-policy objective: SFT fits fixed demonstrations token by token, including targets poorly aligned with the model's pretrained distribution, which can lead to overfitting. 
A recent line of work addresses this issue by assigning larger training weights to tokens better aligned with the current model's predictive distribution, with the intuition that fitting these tokens are less distortive to the model's knowledge and representations. 
However, computing the token weights from the model that is currently fine-tuned entangles token weights with the optimization trajectory, inducing a self-reinforcing loop which concentrates the supervision signal to a narrow set of initially preferred tokens.
To address this, we propose \textbf{PriFT} (\textbf{Pri}or-support guided \textbf{F}ine-\textbf{T}uning), which derives token weights from a frozen pretrained reference to obtain a stable reweighting signal unaffected by fine-tuning. 
This signal estimates \emph{prior support}: the extent to which each target token is supported by the pretrained distribution. 
By reweighting tokens according to their prior support, PriFT limits excessive adaptation to targets requiring substantial deviation from the pretrained distribution and improves SFT generalization.
We introduce two instantiations: \textbf{PriFT-prob} uses pretrained target-token probability, while \textbf{PriFT-mass} selects tokens by cumulative probability mass under the pretrained distribution. 
Across mathematical reasoning, code generation, and medical QA, PriFT achieves state-of-the-art results among SFT baselines, while also providing a stronger initialization for subsequent RL
\footnote{Source code at: \url{https://github.com/wang-kee/PriFT}}. \looseness=-1
\end{abstract}

\section{Introduction}

Post-training is a crucial stage for adapting pretrained large language models (LLMs) to downstream tasks \citep{kumar2025llm},  such as mathematical reasoning \citep{cobbe2021trainingmath, math500} and code generation \citep{wang2021codet5, chen2021evaluatingcode}. 
Supervised fine-tuning (SFT) and reinforcement learning (RL) \citep{ouyang2022rl, sutton1998reinforcement} represent two primary paradigms in post-training. 
SFT is an off-policy method that learns from fixed demonstration data, whereas RL is an on-policy method that optimizes rewards assigned to outputs generated by the model itself. \looseness=-1

Compared with RL, SFT does not require rollout generation during training and provides dense token-level supervision, offering better computational and learning efficiency. 
Moreover, SFT can be applied when no reliable reward model is available, and often serves as the initialization stage for subsequent RL. 
However, prior work has identified a generalization gap between SFT and RL \citep{chu2025sftmem, shenfeld2025rl}. 
As an off-policy method, SFT fits every token in fixed target demonstrations, which can induce overfitting and distort pretrained knowledge when the targets are misaligned with the model's own distribution. In contrast, RL optimizes on samples generated by the model itself, reducing distribution mismatch and better preserving the model's knowledge structure. \looseness=-1

To improve SFT generalization, recent work selectively reweights target tokens according to their alignment with the model's predictive distribution \citep{wu2025dft, diao2026eaft, zhu2025asft, lin2025talr, zhang2026idft, yu2026ranktuner}. 
These methods estimate token-level utility from prediction statistics such as target-token probability or entropy, to modulate each token's contribution and reduce updates from tokens weakly supported by the model's predictive distribution. 
However, they typically compute these statistics from the model currently being fine-tuned, raising an important question: \emph{does the predictive distribution during fine-tuning still reflect the model's pretrained knowledge structure?} 
We show that it does not: the online distribution rapidly drifts from the pretrained distribution, creating a self-reinforcing loop that concentrates the learning signal on initially preferred tokens. 
As a result, the fine-tuned model's distribution no longer faithfully represents the pretrained knowledge distribution, but becomes entangled with optimization dynamics.  \looseness=-1

In this work, we revisit token-reweighted SFT by deriving a cleaner reweighting signal directly from the pretrained model. 
Empirically, we show that computing token weights from the pretrained model consistently improves performance over using the online model being fine-tuned. 
We interpret the pretrained predictive distribution as an estimate of \emph{prior support}: the extent to which each target token is supported by the model's pretrained knowledge and representations before task-specific adaptation. 
Tokens with stronger prior support are better aligned the pretrained model's knowledge structure, thus yielding learning signals that are less likely to distort existing representations.

\begin{figure}[t]
    \centering
    \includegraphics[width=1.0\linewidth]{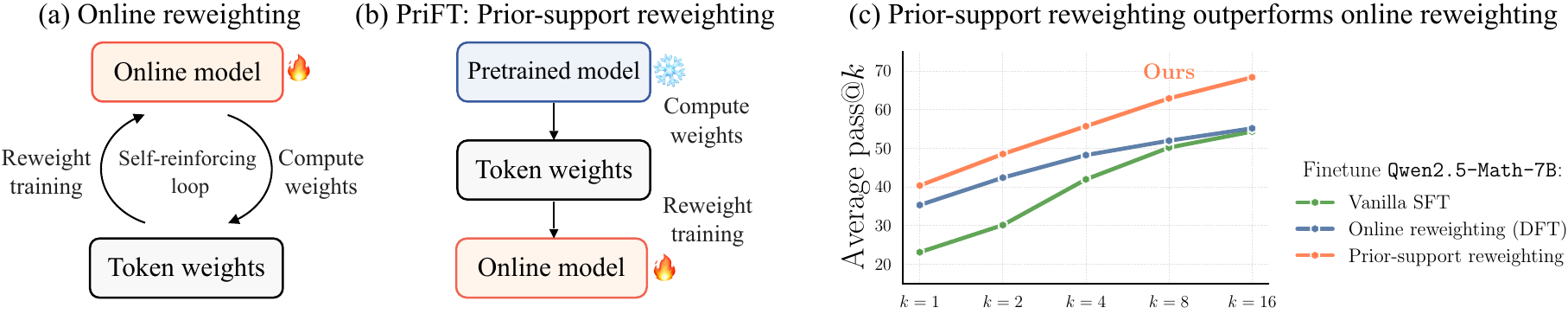}
    \vspace{-1.5em}
    \caption{
    PriFT improves SFT generalization through pretrained-prior guided token reweighting.
    }
    \vspace{-0.95em}
\end{figure}

Based on this principle, we introduce \textbf{PriFT} (\underline{Pri}or-support guided \underline{F}ine-\underline{T}uning), a simple SFT framework that derives token weights from the frozen pretrained model. 
PriFT decouples token-weight estimation from the optimization trajectory and preserves the prior-support signal throughout fine-tuning. 
We study two instantiations: PriFT-prob uses pretrained target-token probability, while PriFT-mass selects tokens by cumulative mass under the pretrained distribution to reduce bias toward easy high-confidence tokens.
Extensive experiments on mathematical reasoning, code generation, and medical question answering show that PriFT achieves state-of-the-art performance among strong token-reweighted SFT baselines. 
Unlike online reweighting methods, PriFT avoids self-reinforcing concentration of supervision signal to a narrow subset of initially preferred tokens, thereby preserving greater sampling diversity and yielding stronger pass@$k$ performance.
\textbf{Notably, simply switching the same probability-based reweighting signal from online model to pretrained reference improves Pass@16 by 14.76 points on \texttt{Qwen2.5-Math-7B}.}
We further show that PriFT provides a better initialization for subsequent RL training, suggesting that pretrained-reference weighting preserves capacity for further on-policy optimization. 
\looseness=-1
Our contributions are fourfold:

\vspace{-0.3em}
\begin{itemize}
    \item We show that the frozen pretrained model give a cleaner token-reweighting signal than the online model, yielding consistent gains across existing token-reweighting methods.
    \item We propose {PriFT}, a prior-support guided framework that computes token weights from a frozen pretrained model, providing a stable reweighting signal throughout fine-tuning.
    \item We demonstrate the strongest performance among token-reweighted SFT methods across mathematical reasoning, code generation, and medical QA, with representative gains of $\bm{3.97}$ Avg@16 and $\bm{8.75}$ Pass@16 points over the strongest prior baseline on \texttt{Qwen2.5-Math-7B}. \looseness=-1
    \item We show that PriFT provides a stronger RL initialization than baseline SFT methods, yielding up to $\bm{9.57}$ points of Avg@16 improvement after RL.
\end{itemize}

\section{Background}

\textbf{Notation and problem formulation. \ }
Let $\mathcal{D}=\{(\bmx^{(n)},\bmy^{(n)})\}_{n=1}^{N}$ denote a dataset of prompt-response pairs. 
For $(\bmx,\bmy)\in\mathcal{D}$, $\bmx$ is the input prompt and $\bmy=(y_1,\dots,y_T)$ is the target response. 
Let $\pi_{\bmth}$ be a language model parameterized by $\bmth$, initialized from a pretrained model $\pi_{\bmth_{\mathrm{pt}}}$. 
SFT minimizes token-level negative log-likelihood:
$
\mathcal{L}_{\textsc{sft}}(\bmth)
=
\mathbb{E}_{(\bmx,\bmy)\sim\mathcal{D}}
\left[
-\sum_{t=1}^{T}
\log \pi_{\bmth}(y_t \mid \bmx, y_{<t})
\right].
$
We refer to the model being fine-tuned as the online model $\pi_{\bmth_{\mathrm{on}}}$, in contrast to the pretrained model $\pi_{\bmth_{\mathrm{pt}}}$. \looseness=-1

Prior work generalizes the standard SFT objective by assigning non-uniform weights to individual tokens during fine-tuning, yielding the weighted SFT objective:
\looseness=-1
\begin{equation}
\mathcal{L}_{\textsc{wsft}}(\bmth)
=
\mathbb{E}_{(\bmx,\bmy)\sim\mathcal{D}}
\left[
-\sum_{t=1}^{T}
m_t \log \pi_{\bmth}(y_t \mid \bmx, y_{<t})
\right],
\label{eq:wsft}
\end{equation}
where $m_t \ge 0$ controls contribution of token $y_t$ to training; setting all $m_t=1$ recovers standard SFT. \looseness=-1

\textbf{Related work on token reweighting with model-prediction statistics. \ }
To improve SFT generalization, recent work derives token weights from the online model's predictive distribution 
$\pion(\cdot \mid \bmx, y_{<t})$ under teacher forcing 
\citep{wu2025dft, diao2026eaft, liu2026profit, lin2025talr, zhu2025asft, zhang2026idft, yu2026ranktuner}. 
This distribution provides a token-level diagnostic of how each supervised target aligns with the current model, and is therefore used to estimate token utility or difficulty. 
Existing methods convert such statistics into token weights to emphasize tokens better aligned with model's distribution and reduce potentially harmful updates. 
Two statistics are commonly used: 
target probability: $p_t=\pion(y_t\mid \bmx,y_{<t})$, which measures the online model's confidence in the target token; and 
prediction entropy: 
$H_t = -\sum_{v\in\mathcal{V}} \pion(v \mid \bmx, y_{<t}) \log \pion(v \mid \bmx, y_{<t})$, 
which measures the uncertainty over the next-token distribution.

\begin{wraptable}{r}{0.31\columnwidth}
\vspace{-18pt}
\centering
\scriptsize
\setlength{\tabcolsep}{4pt}
\renewcommand{\arraystretch}{1.12}
\caption{Summary of baseline token-weighting rules. $\text{sg}(\cdot)$ represents stop gradient. \looseness=-1}
\begin{tabular}{@{}ll@{}}
\toprule
\textbf{Method} & \textbf{Token weight ($m_t$} in \autoref{eq:wsft}) \\
\midrule
SFT 
& $1$ \\
DFT 
& $\operatorname{sg}(p_t)$ \\
ProFit 
& $\mathbf{1}[p_t\ge \tau]$ \\
TALR 
& $\max\{\operatorname{sg}(p_t^{1/\tau}),w_{\min}\}$ \\
EAFT 
& $\widetilde{H}_t = H_t^{\mathrm{top}\text{-}K}/\log K$ \\
ASFT 
& $\operatorname{sg}(p_t)$ with KL anchoring \\
IDFT 
& $p_t^{\exp(-\phi_t)},\ \phi_t=\log p_t+H_t$ \\
\bottomrule
\end{tabular}
\label{tab:baseline_weights}
\vspace{-8pt}
\end{wraptable}

\autoref{tab:baseline_weights} summarizes representative token-reweighted SFT methods. 
Confidence-based methods, including DFT \citep{wu2025dft}, TALR \citep{lin2025talr}, and ProFit \citep{liu2026profit}, use target-token probability to emphasize tokens already supported by the online model. 
DFT directly uses $p_t$ as a soft weight, TALR further applies temperature scaling and lower-bound clipping, and ProFit converts the same signal into a hard selection mask. 
Entropy-based methods use predictive uncertainty: EAFT emphasizes high-entropy tokens \citep{diao2026eaft}, while IDFT combines target probability and entropy through a centered log-likelihood statistic \citep{zhang2026idft}. 
ASFT further adds a KL anchoring term to DFT to constrain the online model toward the pretrained model \citep{zhu2025asft}. \looseness=-1

\section{Pretrained model provides more reliable token-reweighting signals}
\label{sec:pretrained_reference}

Despite the variety of weighting rules, existing methods typically derive token weights from currently fine-tuned model's prediction distribution. This motivates a simple question:
\begin{list}{}{
    \setlength{\leftmargin}{2em}
    \setlength{\rightmargin}{2em}
    \setlength{\topsep}{0pt}
    \setlength{\partopsep}{0pt}
    \setlength{\parsep}{0pt}
    \setlength{\itemsep}{0pt}
}
\item \centering\itshape
Is the online model's predictive distribution the best source for computing token weights?
\end{list}
We argue that it is often not. 
Although online token-level statistics reflect the predictive distribution of the current model, they make token weights dependent on the optimization trajectory. 
As fine-tuning updates the model, the weighting signal becomes a moving target and may no longer faithfully capture the token-level properties it is intended to measure. 
We argue that the pretrained model provides a more reliable source of token statistics, and provides both empirical evidence and analyses. 

\subsection{Empirical motivation}

We begin with a controlled comparison on medical question answering to isolate the effect of the \emph{source of token weights} while keeping the weighting rule fixed. 
Following \citet{zhu2025asft}, we fine-tune \texttt{LLaMA-2-7B}\footnote{\citet{zhu2025asft} use \texttt{LLaMA-2-7B} to reduce potential contamination from prior supervised knowledge.} on 10k MedMCQA \citep{pal2022medmcqa} examples and evaluate on MMLU-Medical \citep{mmlu}, MedQA \citep{medqa}, and MedMCQA test set, reporting average accuracy across the three benchmarks. \looseness=-1

We consider DFT-style target-probability reweighting \citep{wu2025dft}, top-50\% and bottom-50\% token selection by target probability, and top-50\% and bottom-50\% token selection by predictive entropy. 
For each rule, we compare token reweighting using predictions from the online model versus the frozen pretrained model.

\begin{table}[h]
\centering
\small
\setlength{\tabcolsep}{5pt}
\caption{\textbf{Pretrained-model reweighting consistently improve online-model reweighting. } %
}
\vspace{-5pt}
\resizebox{0.9\linewidth}{!}{%
\begin{tabular}{
c
>{\columncolor{gray!10}}c
>{\columncolor{gray!10}}c
ccccc
}
\toprule
Token reweighting model & Original & SFT & $p_t$ rw. & $p_t$ top-50 & $p_t$ bot.-50 & $H_t$ top-50 & $H_t$ bot.-50  \\
\midrule
Online model     & 31.40 & 33.37 & 29.59 & 27.32 & 33.78 & 31.04 & 24.27 \\
Pretrained model & 31.40 & 33.37 & \textbf{35.81} & \textbf{30.59} & \textbf{35.68} & \textbf{33.91} & \textbf{27.37}  \\
\bottomrule
\end{tabular}%
}
\label{tab:motivation_pretrained_stats}
\vspace{-0.8em}
\end{table}

\autoref{tab:motivation_pretrained_stats} shows that reweighting with pretrained model consistently outperform using online model across all tested rules, including confidence-based reweighting and selection, and entropy-based selection \footnote{These experiments use a low-data regime to enable a controlled comparison of token-statistics sources. In this setting, token reweighting can underperform standard SFT, as also observed by \citet{zhu2025asft}.}. 
This indicates that the token reweighting source is a key design choice, and a pretrained model can provide a more effective weighting signal than the evolving online model in this empirical setting. \looseness=-1

\subsection{Analysis: Why the pretrained reference improves token reweighting}
\label{sec:analysis}

Having established the empirical benefit, we next analyze why a clean pretrained model provides a better signal for token reweighting. Let
$
p^{\mathrm{pt}}_t=\pi_{\bmth_{\mathrm{pt}}}(y_t\mid\bmx,y_{<t})
$
and
$
p^{\mathrm{on}}_t=\pi_{\bmth_{\mathrm{on}}}(y_t\mid\bmx,y_{<t})
$
denote the target probability under the pretrained reference and the online model, respectively. We analyze a model fine-tuned on 10k MedMCQA examples with DFT \citep{wu2025dft}, a representative online reweighting method that reweights each token's contribution by $p^{\mathrm{on}}_t$. We then examine three factors that explain why signals derived from the online model are less reliable. \looseness=-1

\begin{figure}
    \centering
    \includegraphics[width=1.0\linewidth]{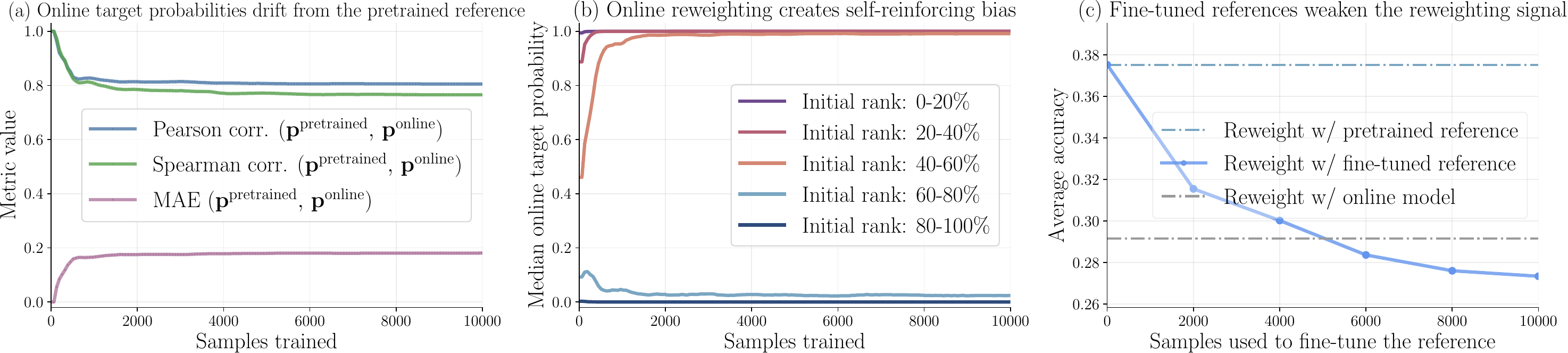}
    \caption{\textbf{A clean pretrained reference provides more reliable token-level reweighting signals.}
    (a) Online target probabilities drift rapidly from pretrained probabilities: correlations decrease early in training.
    (b) Online reweighting amplifies initial token-rank bias: tokens with high pretrained support are pushed toward near-deterministic probabilities, while low-ranked tokens remain weakly supported.
    (c) Reweighting signals derived from partially fine-tuned references become progressively less effective, showing the importance of preserving an uncontaminated pretrained reference. \looseness=-1}
    \label{fig:pt_reference}
    \vspace{-0.8em}
\end{figure}

\textbf{Online reweighting signal drifts rapidly during fine-tuning. \ }
When token statistics are computed from the online model, the weighting signal changes together with the model being fine-tuned. To quantify this drift, we compare the pretrained target probability $\bm{p}^{\mathrm{pt}}$ with the online probability $\bm{p}^{\mathrm{on}}$ during fine-tuning on response tokens from 64 held-out examples, using Pearson correlation, Spearman correlation, and mean absolute error.
\autoref{fig:pt_reference}(a) shows that the drift occurs mostly at the beginning of fine-tuning. 
Both correlations quickly drop to around $0.8$, while the mean absolute error rises to about $0.18$. Thus, online-derived signals rapidly depart from the initial pretrained reference and remain shifted throughout fine-tuning, resulting in non-stable token weights. \looseness=-1

\textbf{Online reweighting creates a self-reinforcing bias. \ }
This drift is not merely random: it induces a self-reinforcing bias in the token weights. Tokens that are initially favored by the pretrained model receive larger online probabilities, and therefore larger weights and updates, while initially disfavored tokens receive progressively weaker supervision.
To visualize this effect, we group response tokens from the 64 held-out examples into five equal-sized bins according to their initial rank under $p_t^{\mathrm{pt}}$, and track the median value of $p_t^{\mathrm{on}}$ in each bin during fine-tuning. As shown in \autoref{fig:pt_reference}(b), tokens in the initially favored bins are rapidly pushed toward near-deterministic target probabilities, whereas the lowest-ranked bins are pushed toward zero. This reveals a rich-get-richer dynamic: online reweighting amplifies the model's initial preference and concentrates supervision on a subset of preferred tokens \footnote{We provide an example of the self-reinforcing bias of online reweighting in \autoref{fig:token_prob_heatmap} in the appendix.}. \looseness=-1

\textbf{Being frozen is not enough: the pretrained reference must remain uncontaminated. \ }
One possible explanation is that the pretrained model benefits token reweighting simply because it provides a fixed reference, in which case any frozen reference should be similarly effective.
We test this by constructing frozen reference models from different stages of fine-tuning. Specifically, we fine-tune reference models with DFT on varying numbers of training samples, freeze each checkpoint, and use it to provide reweighting signal for the training of a new downstream model initialized from the original pretrained checkpoint on a separate 10k-sample training set. 
As shown in \autoref{fig:pt_reference}(c), the best performance of the downstream model is achieved using original pretrained model as reference. Fine-tuning the reference on as few as 2k samples already causes a substantial drop, and performance continues to decline with further fine-tuning, eventually falling below online-model reweighting. These results indicate that freezing alone is insufficient: preserving the pretrained distribution is critical, as fine-tuning progressively degrades the usefulness of the reference token statistics for reweighting. \looseness=-1

For comparison, we repeat the same analysis for a model trained with standard SFT in \autoref{fig:pt_reference_app} in appendix. Compared with DFT, SFT exhibits smaller drift from the pretrained reference, weaker self-reinforcing bias, and a milder performance drop when providing reweighting signals. The comparison suggests that these effects are not only caused by fine-tuning, 
but amplified by online reweighting.\looseness=-1

\subsection{Interpretation: pretrained model provides an estimate of prior support}

The results suggest that the pretrained model is not merely a fixed source of token reweighting. 
Rather, the pretrained target-token probability provides an estimate of \emph{prior support}: how strongly each target token is supported by the model's pretrained representations before task-specific adaptation. \looseness=-1

We interpret prior support as a proxy for adaptation cost: the negative log-probability of a target token under the pretrained model corresponds to the KL divergence from the one-hot target distribution to the pretrained distribution. Thus, tokens with lower prior support exhibit greater initial distribution mismatch and may require more substantial adaptation.
Crucially, the prior support signal must remain pretrained, as fine-tuning entangles it with optimization dynamics and compromises its reliability.\looseness=-1

\section{PriFT: Prior-support Guided Fine-tuning}
\label{sec:PriFT}

We propose \textbf{PriFT} (\underline{Pri}or-support guided \underline{F}ine-\underline{T}uning), a framework that constructs token weights from the prior support from pretrained model, decoupling reweighting from optimization trajectory \footnote{The pretrained model refers to the checkpoint before SFT in this paper, not necessarily the checkpoint after pretraining. \looseness=-1}.

\textbf{PriFT framework. }
For each training example $(\bmx,\bmy)$ and target token $y_t$, PriFT assigns a token weight using the pretrained predictive distribution:
$
m_t
=
u\!\left(\pi_{\bmth_{\mathrm{pt}}}(\cdot \mid \bmx, y_{<t}),\, y_t\right),
$
where $u(\cdot)$ is a token-utility function that maps the pretrained distribution and the target token to a nonnegative weight $m_t$.
The online model is then trained with the weighted SFT objective
$
\mathcal{L}_{\textsc{PriFT}}(\bmth_{\mathrm{on}})
=
-\sum_{t=1}^{T}
u\!\left(\pi_{\bmth_{\mathrm{pt}}}(\cdot \mid \bmx, y_{<t}),\, y_t\right)
\log \pi_{\bmth_{\mathrm{on}}}(y_t \mid \bmx, y_{<t}).
$

By reweighting token contributions according to their prior support, PriFT encourages \textit{conservative adaptation}: it fits the target data while limiting excessive deviation from the pretrained distribution. This preserves task adaptation while reducing the influence of tokens that are poorly aligned with the pretrained model and may distort pretrained knowledge.
PriFT is a general reweighting framework, not a single weighting rule. 
We study two instantiations: \textbf{PriFT-prob}, based on pretrained target-token probability, and \textbf{PriFT-mass}, based on cumulative-mass support under the pretrained distribution. \looseness=-1 %

\textbf{PriFT-prob: Probability-based reweighting from the pretrained reference. \ }
PriFT-prob is the most direct instantiation of PriFT. 
It uses the probability-based utility of DFT \citep{wu2025dft}, but computes the target-token probability from the frozen pretrained reference instead. 
For each target token $y_t$, we define
$
m_t
=
\pi_{\bmth_{\mathrm{pt}}}(y_t \mid \bmx, y_{<t}),
$
which gives
{\setlength{\abovedisplayskip}{2pt}
\setlength{\belowdisplayskip}{2pt}
\begin{equation}
\label{eq:method_prob}
\mathcal{L}_{\textsc{PriFT-prob}}(\bmth_{\mathrm{on}})
=
-\sum_{t=1}^{T}
\pi_{\bmth_{\mathrm{pt}}}(y_t \mid \bmx, y_{<t})
\log \pi_{\bmth_{\mathrm{on}}}(y_t \mid \bmx, y_{<t}).
\end{equation}}
PriFT-prob scales each token's contribution according to its pretrained support, assigning stronger supervision to tokens that are more plausible under the pretrained distribution while downweighting those with weaker support. Because the reweighting signal remains fixed throughout fine-tuning, it avoids the self-reinforcing dynamics of online probability reweighting discussed in Section \ref{sec:analysis}.

PriFT-prob works best when fine-tuning targets are well covered by the pretrained distribution, so that pretrained probability provides a reliable learning signal. However, raw probability can overemphasize easy high-confidence tokens. In knowledge-intensive or domain-specific settings, important tokens may have low initial probability due to limited domain coverage, causing direct probability weighting to suppress updates needed for adaptation. This motivates PriFT-mass, which keeps the same prior-support principle but replaces raw probability with a cumulative support measure. \looseness=-1

\begin{wrapfigure}{r}{0.27\linewidth}
    \centering
    \vspace{-1.2em}
    \includegraphics[width=1.0\linewidth]{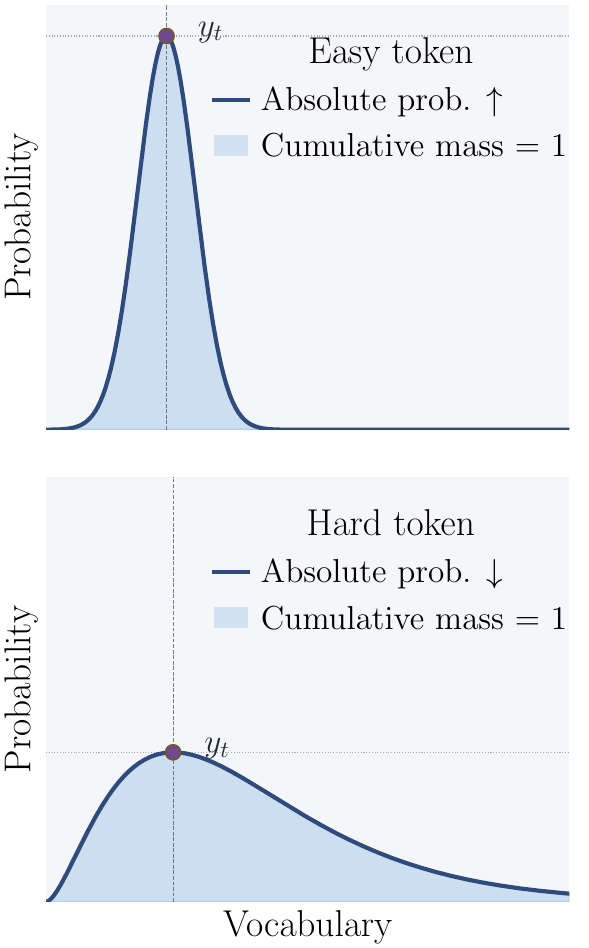}
    \caption{
    Cumulative-mass support reduces the bias toward easy tokens with sharp distributions.
    }
    \label{fig:mass_illustration}
    \vspace{-1.2em}
\end{wrapfigure}

\textbf{PriFT-mass: Cumulative-mass selection from the pretrained reference. \ }
\label{sec:PriFT_mass}
Raw target probability is biased toward easy positions where the pretrained distribution is sharp. 
Even when an easy token and a harder token are both well ranked under the pretrained model, the easy token can receive a much larger absolute probability simply because its local distribution is more concentrated. 
As a result, PriFT-prob may overemphasize high-confidence tokens, including trivial continuations such as function words or formatting tokens, while underweighting harder tokens that encode task-specific knowledge or key reasoning steps. 
This motivates a utility that measures target-token support relative to alternative candidates under the same pretrained distribution, rather than by absolute probability alone. \looseness=-1

To this end, we introduce \textbf{PriFT-mass}, which instantiates PriFT with a cumulative-mass utility. Let $
p_t^{\mathrm{pt}}(v)=\pi_{\bmth_{\mathrm{pt}}}(v \mid \bmx, y_{<t})
$ denote the pretrained reference distribution at token position $t$. We define
$
u_t^{(\textsc{mass})}
=
\sum_{v:\,p_t^{\mathrm{pt}}(v)\le p_t^{\mathrm{pt}}(y_t)}
p_t^{\mathrm{pt}}(v).
$
This score measures the probability mass assigned to tokens no more likely than the target token under the pretrained distribution. 
Thus, a target token receives a large score when it is supported above a substantial fraction of the reference probability mass, even if its raw probability is moderate. 
As illustrated in \autoref{fig:mass_illustration}, a top-ranked target token receives a score of $1$ regardless of whether the distribution is sharp or flat. 
PriFT-mass can therefore retain informative tokens from harder contexts that PriFT-prob may downweight.
We convert this score into a binary token-selection mask with a fixed threshold of $0.5$:{\setlength{\abovedisplayskip}{2pt}
\setlength{\belowdisplayskip}{2pt}\begin{equation}
\mathcal{L}_{\textsc{PriFT-mass}}(\bmth_{\mathrm{on}})
=
-\sum_{t=1}^{T}
m_t \log \pi_{\bmth_{\mathrm{on}}}(y_t \mid \bmx, y_{<t}),
\quad
m_t
=
\mathbf{1}\!\left[
u_t^{(\textsc{mass})}
\ge
0.5
\right].
\label{eq:ft_mass}
\end{equation}}
Since $u_t^{(\textsc{mass})}$ is a percentile-based support score, the threshold $0.5$ keeps tokens supported above at least half of the pretrained probability mass, filtering tokens in the lower-support half. Unlike PriFT-prob, PriFT-mass uses the pretrained reference only for token selection; all selected tokens then receive equal training weight regardless of their prediction difficulty. \looseness=-1 

\textbf{Computational overhead. \ } PriFT introduces limited additional computation. 
Let $F$ denote the FLOPs of one forward pass over the training set. 
Since one backward pass costs approximately $2F$, standard SFT trained for $E$ epochs requires about $3EF$ FLOPs. 
PriFT adds a single frozen forward pass of the pretrained reference model to comput token statistics, giving a total cost of approximately $F+3EF$ and a relative overhead of $1/(3E)$. 
For example, when training for 2 epochs, this corresponds to $16.7\%$ additional computation.

This extra cost can be incurred offline \textit{before} fine-tuning. 
During training, PriFT can use cached token weights or masks and follows the same forward--backward optimization procedure as standard weighted SFT. 
Therefore, PriFT does not necessarily require loading an additional reference model during training, and does not introduce extra GPU memory overhead for maintaining a second model. 
In contrast, KL-based regularization methods between the fine-tuned and pretrained models generally require access to the reference distribution during optimization \citep{zhu2025asft}, as 
caching such distributions would require storing full-vocabulary logits or probabilities for every token, whereas PriFT is much cheaper to cache by storing only one scalar weight or binary mask per target token. \looseness=-1

\section{Experiments}

\subsection{Experimental setup}

We evaluate \methodname across mathematical reasoning, code generation and medical question answering \footnote{We provide the experimental setup and results of medical question answering results in Appendix~\ref{app:medical_results}.}. Furthermore, we evaluate using \methodname as an initialization for subsequent RL stage. \looseness=-1

\textbf{Mathematical reasoning. \ }
For mathematical reasoning, we follow the setup from \citet{wu2025dft}, reporting main results on \texttt{Qwen2.5-Math-7B} and \texttt{Qwen3-8B-Base} \citep{yang2025qwen3}.\footnote{Additional results on \texttt{Qwen2.5-1.5B}, \texttt{Qwen2.5-Math-1.5B}, \texttt{Qwen2.5-Instruct-1.5B}, \texttt{Qwen2.5-14B}, and \texttt{DeepSeekMath-7B} are provided in the appendix.} 
Each model is fine-tuned for one epoch on 100k randomly sampled examples from \texttt{NuminaMath-CoT} \citep{li2024numinamath}. 
We evaluate on \texttt{math\_oai} \citep{math500}, \texttt{minerva\_math} \citep{mineramath}, \texttt{olympiadbench} \citep{he2024olympiadbench}, \texttt{aime24}, and \texttt{amc23}, using the default chat template with chain-of-thought prompting. 
Results are aggregated over 16 stochastic decoding runs: {Avg@16} denotes the mean accuracy across the 16 runs, and {Pass@16} denotes the fraction of examples solved by at least one sampled response. We provide full training and evaluation details in Appendix \ref{app:training_details}.

\textbf{Code generation. \ }
For code generation, we fine-tune \texttt{Qwen2.5-Coder-3B} \citep{hui2024qwen25coder} and \texttt{Qwen3-4B-Instruct} \citep{yang2025qwen3} for one epoch on \texttt{Tulu-3-SFT-Personas-Code} \citep{lambert2024tulu3}, containing 34,999 examples focused on diverse Python coding questions. We evaluate on four benchmarks: \texttt{HumanEval+} \citep{chen2021evaluatingcode,evalplus}, \texttt{MBPP+} \citep{austin2021mbpp,evalplus}, and two recent contest splits from \texttt{LiveCodeBench} \citep{jain2024livecodebench} (v5 and v6). We report pass@1 accuracy on each benchmark.

\textbf{Baselines. \ }
We compare PriFT with standard SFT and several token-reweighted SFT methods. DFT \citep{wu2025dft} and TALR \citep{lin2025talr} derive token weights from the target-token probability, while ASFT \citep{zhu2025asft} extends DFT with a KL regularization term to the pretrained model. IDFT \citep{zhang2026idft} further emphasizes tokens with higher signal-to-noise ratios based on centered log-likelihood of the target token. EAFT \citep{diao2026eaft} instead uses predictive entropy to restrict training to high-entropy tokens. \looseness=-1

\begin{table*}[t]
\caption{Mathematical reasoning performance across five benchmarks with two different models. PriFT achieves the strongest overall performance on both models. \looseness=-1}
\label{tab:main_math_results}
\vspace{0.4em}
\centering
\small
\setlength{\tabcolsep}{4.5pt}
\renewcommand{\arraystretch}{1.12}
\resizebox{\textwidth}{!}{%
\begin{tabular}{clcccccccccccc}
\toprule
\multirow{2}{*}{\textbf{Model}} & \multirow{2}{*}{\textbf{Method}}
& \multicolumn{2}{c}{\textbf{MATH-OAI}}
& \multicolumn{2}{c}{\textbf{Minerva Math}}
& \multicolumn{2}{c}{\textbf{OlympiadBench}}
& \multicolumn{2}{c}{\textbf{AIME24}}
& \multicolumn{2}{c}{\textbf{AMC23}}
& \multicolumn{2}{c}{\textbf{Average}} \\
\cmidrule(lr){3-4}
\cmidrule(lr){5-6}
\cmidrule(lr){7-8}
\cmidrule(lr){9-10}
\cmidrule(lr){11-12}
\cmidrule(lr){13-14}
& & Avg@16 & P@16
  & Avg@16 & P@16
  & Avg@16 & P@16
  & Avg@16 & P@16
  & Avg@16 & P@16
  & Avg@16 & P@16 \\
\midrule

\multirow{8}{*}{\makecell[c]{Qwen2.5-Math\\(7B)}}
& Original & 40.53 & 89.80 & 12.99 & 49.63 & 17.66 & 58.67 & 8.74 & 40.00 & 24.84 & \underline{82.50} & 20.95 & 64.12 \\
\cmidrule(lr){2-14}
& SFT      & 54.17 & 89.00 & 17.26 & 50.74 & 18.79 & 54.52 & 2.49 & 13.33 & 25.00 & 70.00 & 23.54 & 55.52 \\
& DFT      & 68.55 & 84.80 & 27.55 & 42.65 & 34.50 & 56.00 & 7.93 & 13.33 & 40.94 & 72.50 & 35.89 & 53.86 \\
& EAFT     & 53.11 & 87.80 & 18.84 & \underline{52.94} & 18.85 & 54.67 & 2.29 & 13.33 & 22.19 & 75.00 & 23.06 & 56.75 \\
& IDFT     & 66.76 & 82.40 & \underline{27.04} & 41.18 & 31.71 & 49.78 & 6.47 & 20.00 & 45.31 & 77.50 & 35.46 & 54.17 \\
& TALR     & 69.70 & 88.20 & \textbf{29.12} & 45.59 & 34.71 & 59.85 & 7.93 & 16.67 & 42.03 & 80.00 & 36.70 & 58.06 \\
& ASFT     & 68.38 & \underline{91.80} & 24.82 & 51.84 & 33.95 & \underline{63.11} & 7.93 & 20.00 & 42.50 & 77.50 & 35.90 & 60.85 \\
\rowcolor{blue!5}
& \methodname-prob & \underline{73.12} & \underline{91.80} & 26.40 & 52.21 & \textbf{36.14} & \textbf{65.78} & \underline{12.72} & \underline{43.33} & \underline{49.84} & \textbf{90.00} & \underline{39.65} & \underline{68.62} \\
\rowcolor{blue!5}
\rowcolor{blue!5}
& \methodname-mass & \textbf{73.83} & \textbf{92.20} & 26.71 & \textbf{54.04} & \underline{35.75} & 61.78 & \textbf{13.34} & \textbf{50.00} & \textbf{53.75} & \textbf{90.00} & \textbf{40.67} & \textbf{69.60} \\
\midrule

\multirow{8}{*}{\makecell[c]{Qwen3-Base\\(8B)}}
& Original & 57.93 & \textbf{93.20} & 20.61 & \underline{54.41} & 27.23 & \textbf{66.07} & 8.13 & \textbf{33.33} & 35.47 & \textbf{85.00} & 29.87 & \textbf{66.40} \\
\cmidrule(lr){2-14}
& SFT      & 51.61 & 88.60 & 20.82 & \textbf{56.62} & 19.00 & 54.07 & 1.86 & 13.33 & 22.81 & 62.50 & 23.22 & 55.03 \\
& DFT      & 69.01 & 85.20 & \underline{29.38} & 49.63 & 33.34 & 58.07 & 6.46 & 16.67 & 43.59 & 77.50 & 36.36 & 57.41 \\
& EAFT     & 53.10 & 88.00 & 19.40 & \underline{55.15} & 18.90 & 53.93 & 2.30 & 13.33 & 24.20 & 70.00 & 23.58 & 56.08 \\
& IDFT     & \underline{69.70} & 82.80 & 24.20 & 37.87 & 33.20 & 49.33 & 7.50 & 10.00 & 43.00 & 67.50 & 35.51 & 49.50 \\
& TALR     & 68.60 & 88.80 & 28.50 & 49.26 & 31.70 & 58.67 & 5.60 & 16.67 & 40.90 & \underline{82.50} & 35.07 & 59.18 \\
& ASFT     & 67.80 & \underline{91.40} & 28.07 & 51.84 & 32.41 & 63.41 & 7.72 & 20.00 & 39.06 & \textbf{85.00} & 35.01 & 62.33 \\
\rowcolor{blue!5}
& \methodname-prob & 68.65 & \underline{91.40} & 28.03 & \underline{54.41} & \underline{34.20} & 63.85 & \underline{8.33} & \underline{26.67} & \underline{44.06} & \underline{82.50} & \underline{36.65} & 63.77 \\
\rowcolor{blue!5}
& \methodname-mass & \textbf{70.99} & 90.80 & \textbf{30.43} & 54.78 & \textbf{36.36} & \underline{64.00} & \textbf{11.24} & \textbf{33.33} & \textbf{47.03} & \textbf{85.00} & \textbf{39.21} & \underline{65.58} \\
\bottomrule
\end{tabular}%
}
\vspace{-1em}
\end{table*}

\subsection{Main results on supervised fine-tuning}

\paragraph{Mathematical reasoning.}
\autoref{tab:main_math_results} reports results on five mathematical reasoning benchmarks. 
Across both backbones, PriFT-mass achieves the strongest aggregate performance among fine-tuning methods, closely followed by PriFT-prob.
On \texttt{Qwen2.5-Math-7B}, PriFT-mass reaches $40.67$ Avg@16 and $69.60$ Pass@16, improving over the strongest prior fine-tuning baseline by $3.97$ and $8.75$ points, respectively. 
On \texttt{Qwen3-8B-Base}, PriFT-mass obtains $39.21$ Avg@16 and $65.58$ Pass@16, outperforming the strongest prior baseline by $2.85$ and $3.25$ points. 

A notable pattern is that online-reweighting methods often improve Avg@16 at the cost of Pass@16. 
On \texttt{Qwen2.5-Math-7B}, DFT improves Avg@16 over SFT by $12.35$ points, but reduces Pass@16 by $1.66$ points; in contrast, PriFT-mass improves Avg@16 by $17.13$ points and Pass@16 by $14.08$ points. 
On \texttt{Qwen3-8B-Base}, DFT improves Avg@16 over SFT by $13.14$ points but gains only $2.38$ points in Pass@16, while PriFT-mass improves Avg@16 by $15.99$ points and Pass@16 by $10.55$ points. \looseness=-1

The comparison with DFT directly isolates the effect of the reweighting-signal source. 
DFT and PriFT-prob share the same probability-based weighting form, but differ only in the source of the probability: DFT uses the evolving online model, whereas PriFT-prob uses the frozen pretrained model. 
This change in the reweighting signal improves Pass@16 over DFT by $14.76$ points on \texttt{Qwen2.5-Math-7B} and $6.36$ points on \texttt{Qwen3-8B-Base}.
This suggests that online reweighting can concentrate probability mass on a narrower set of preferred reasoning paths, whereas a fixed pretrained-reference signal mitigates this trade-off and better preserves diverse successful trajectories.
Additionally, we show that the gains are robust across independent random seeds (Appendix \ref{app:seed_robustness}), and remain effective after normalizing token weights (Appendix \ref{app:weight_normalization}).
\begin{table}[t]
    \centering
    \begin{minipage}[t]{0.49\textwidth}
        \centering

\caption{Code generation task performance using \texttt{Qwen3-4B-Instruct}. We report pass@1 accuracy on each benchmark and the average.\looseness=-1}
\resizebox{\linewidth}{!}{%
\begin{tabular}{lccccc}
\toprule
\textbf{Method}
& \textbf{HumanEval+} & \textbf{MBPP+} & \textbf{LCB\,v5} & \textbf{LCB\,v6} & \textbf{Avg} \\
\midrule
Original
& \textbf{82.93} & \underline{65.61} & 30.45 & 28.34 & 51.83 \\
SFT
& 68.29 & 63.23 & 22.39 & 21.52 & 43.86 \\
DFT
& 60.37 & 60.85 & 24.32 & 23.22 & 42.19 \\
IDFT
& 65.85 & 61.90 & 26.14 & 24.74 & 44.66 \\
EAFT
& 68.29 & 63.76 & 23.86 & 23.13 & 44.76 \\
TALR
& 60.98 & 65.34 & 25.00 & 23.60 & 43.73 \\
ASFT
& 70.73 & 64.29 & 28.64 & 26.73 & 47.60 \\
\rowcolor{blue!5}
\methodname-prob
& 76.83 & \textbf{67.99} & \underline{39.32} & \underline{36.87} & \textbf{55.25} \\
\rowcolor{blue!5}
\methodname-mass
& \underline{78.66} & 64.81 & \textbf{39.89} & \textbf{37.25} & \underline{55.15} \\
\bottomrule
\end{tabular}
}
\label{tab:main_code_results}

    \end{minipage}
    \hfill
    \begin{minipage}[t]{0.485\textwidth}
        \centering
          \caption{RL performance from different initialized SFT checkpoints, including relative improvement compared to prior-RL. \looseness=-1
  }
  \vspace{-0.05em}
  \centering
  \resizebox{\linewidth}{!}{%
  \small
  \setlength{\tabcolsep}{8pt}
  \renewcommand{\arraystretch}{1.0}
  \begin{tabular}{clcc}
  \toprule
  \textbf{Model} & \textbf{Init.} & \textbf{Avg@16 \textcolor{ForestGreen}{($\Delta$)}} & \textbf{Pass@16 \textcolor{ForestGreen}{($\Delta$)}} \\
  \midrule
  \multirow{4}{*}{\shortstack[c]{Qwen2.5-Math\\1.5B}}
  & SFT        & 32.22 \textcolor{ForestGreen}{(+14.4)} & 61.58 \textcolor{ForestGreen}{(+11.3)} \\
  & DFT        & 35.22 \textcolor{ForestGreen}{(+3.81)}  & 59.13 \textcolor{ForestGreen}{(+4.89)}  \\
  & \cellcolor{blue!5}PriFT-prob
  & \cellcolor{blue!5}\underline{38.64} \textcolor{ForestGreen}{(+7.26)}
  & \cellcolor{blue!5}\underline{65.96} \textcolor{ForestGreen}{(+2.25)} \\
  & \cellcolor{blue!5}PriFT-mass
  & \cellcolor{blue!5}\textbf{38.96} \textcolor{ForestGreen}{(+6.26)}
  & \cellcolor{blue!5}\textbf{67.23} \textcolor{ForestGreen}{(+2.71)} \\
  \midrule
  \multirow{4}{*}{\shortstack[c]{Qwen3-8B\\Base}}
  & SFT        & 42.67 \textcolor{ForestGreen}{(+19.5)} & \textbf{70.24} \textcolor{ForestGreen}{(+15.2)} \\
  & DFT        & 39.63 \textcolor{ForestGreen}{(+3.28)}  & 59.40 \textcolor{ForestGreen}{(+1.99)} \\
  & \cellcolor{blue!5}PriFT-prob
  & \cellcolor{blue!5}\underline{46.23} \textcolor{ForestGreen}{(+9.57)}
  & \cellcolor{blue!5}\underline{70.16} \textcolor{ForestGreen}{(+6.40)} \\
  & \cellcolor{blue!5}PriFT-mass
  & \cellcolor{blue!5}\textbf{47.65} \textcolor{ForestGreen}{(+8.44)}
  & \cellcolor{blue!5}67.58 \textcolor{ForestGreen}{(+1.99)} \\
  \bottomrule
  \end{tabular}%
  }
  \label{tab:rl_initialization}

    \end{minipage}
\end{table}

\textbf{Code generation. \ }
\autoref{tab:main_code_results} reports code generation results on \texttt{Qwen3-4B-Instruct}, with results on \texttt{Qwen2.5-Coder-3B} deferred to Appendix~\ref{app:code_generation}. 
Both PriFT variants outperform standard SFT and prior token-reweighted baselines on average, with the largest gains on the more recent
LiveCodeBench splits. 
PriFT-prob achieves the best average accuracy of $55.25$, improving over the strongest prior baseline by $7.65$ points, while PriFT-mass reaches a comparable $55.15$. 

A notable pattern is that fine-tuning often degrades performance on HumanEval+ and MBPP+, where the original model already performs strongly. 
No fine-tuning method improves over the original checkpoint on HumanEval+, but PriFT incurs the smallest degradation: PriFT-mass drops by only $4.27$ points, compared with $14.64$ for SFT. 
At the same time, PriFT improves substantially on the more recent LiveCodeBench splits: PriFT-mass improves over the closest fine-tuning baseline by $11.25$ points on LCB v5 and $10.52$ points on LCB v6. 
This suggests that pretrained-reference weighting better preserves existing capabilities while still enabling adaptation to harder coding tasks. 

\textbf{Medical QA. \ }
\autoref{tab:medical_benchmark} in Appendix~\ref{app:medical_results} reports medical QA results with \texttt{LLaMA-2-7B}. PriFT-mass, together with ASFT, achieves the strongest overall performance, obtaining the best results on MMLU and MedMCQA. PriFT-prob also performs strongly, outperforming the remaining token-reweighted SFT baselines overall. These results show that PriFT generalizes beyond math and code to knowledge-intensive domains such as medical question answering. \looseness=-1

\subsection{PriFT provides a better initialization for RL}
\label{sec:PriFT_rl}

Across diverse evaluation domains, PriFT consistently improves supervised fine-tuning performance, we next evaluate how different SFT methods affect subsequent RL performance. Performing RL after SFT is a standard recipe for further improving model capabilities, especially on reasoning tasks. We compare standard SFT, DFT, and PriFT-variants as initializations and apply DAPO~\citep{yu2025dapo} to each post-SFT checkpoint \footnote{Due to the large computational cost for RL training, we considered only two representative SFT baselines.}, using \texttt{Qwen2.5-Math-1.5B} and \texttt{Qwen3-8B-Base} as backbones. For each backbone, all initializations share the same RL configuration provided in Appendix~\ref{app:rl_setup}. \looseness=-1

\autoref{tab:rl_initialization} shows that PriFT achieves the strongest post-RL performance overall. On \texttt{Qwen2.5-Math-1.5B}, PriFT-mass reaches $38.96$ Avg@16 and $67.23$ Pass@16, improving over the strongest baseline by $3.74$ and $5.65$ points, respectively. On \texttt{Qwen3-8B-Base}, PriFT-mass achieves the best Avg@16 of $47.65$, improving over the strongest baseline by $4.98$ points, while PriFT-prob matches the strongest Pass@16 performance with $70.16$.  The results reveal different failure modes of the baselines: SFT benefits substantially from RL but delivers weaker supervised checkpoint, while DFT yields a stronger SFT checkpoint but gains less from RL, especially on Pass@16. This is consistent with our earlier finding that online reweighting can concentrate probability mass on a narrower set of reasoning trajectories, reducing the headroom for RL to discover additional correct solutions. In contrast, PriFT provides strong supervised checkpoints that remain amenable to further on-policy optimization.  In the next section, we show that this benefit is associated with PriFT preserving greater sampling diversity than online-reweighting methods. \looseness=-1

\section{Analyses and Discussions}

We compare \methodname with DFT, a representative online token-reweighting method, in both sampling diversity and distribution shift during fine-tuning, and present ablations to justify the design choices. \looseness=-1

\subsection{\methodname preserves more sampling diversity}

\begin{figure}[t]
    \centering
    \vspace{-0em}
    \includegraphics[width=1.0\linewidth]{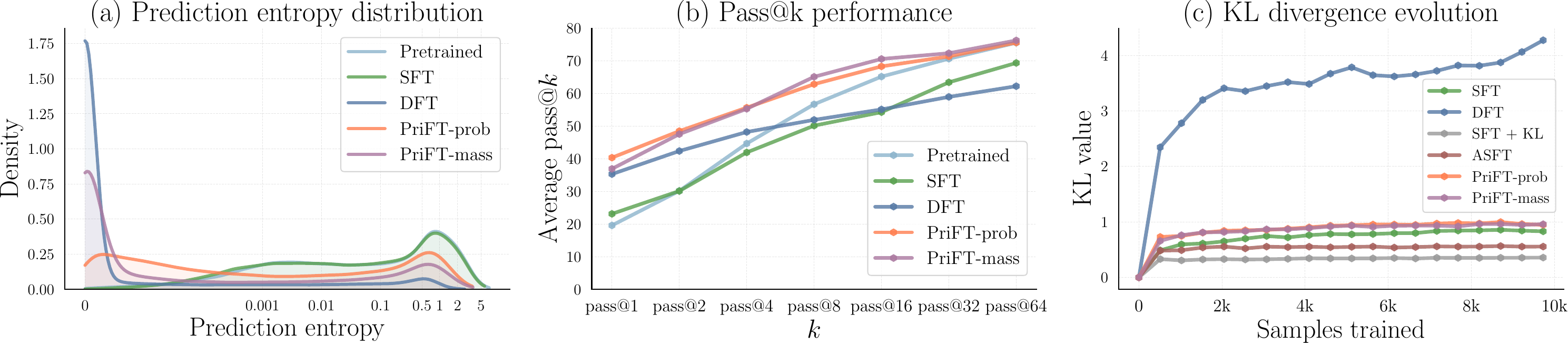}
    \caption{
    (a) PriFT preserves broader token-level entropy distributions than DFT. 
    (b) PriFT achieves strong pass@$k$ performance across different sampling budgets.
    (c) PriFT induces substantially smaller KL divergence from the pretrained model than DFT during training.
    }
    \label{fig:pred_entropy}
    \vspace{-1em}
\end{figure}

We show PriFT preserves higher sampling diversity. This is important for sampling-based reasoning evaluation and subsequent RL, where maintaining multiple plausible trajectories can provide additional optimization headroom.
We evaluate the prediction entropy and pass@$k$ performance for different fine-tuning methods on \texttt{Qwen2.5-Math-7B}. \looseness=-1

\textbf{Prediction entropy.}
We compare token-level prediction entropy on response tokens from 64 held-out samples after fine-tuning. 
As shown in \autoref{fig:pred_entropy}(a), DFT places a large fraction of tokens near zero entropy, indicating that online probability reweighting makes the model overly deterministic through its self-reinforcing bias. 
In contrast, PriFT variants maintain broader entropy distributions, suggesting that pretrained-reference weighting better preserves uncertainty in the predictive distribution. \looseness=-1

\textbf{Pass@$k$ with increasing $k$.}
We further evaluate average pass@$k$ performance as the number of sampled responses $k$ increases. 
\autoref{fig:pred_entropy}(b) shows that although DFT achieves strong pass@$k$ at small values of $k$, its performance saturates more quickly than the other methods. 
In comparison, PriFT-prob and PriFT-mass maintain strong pass@$k$ performance across sampling budgets from $k=1$ to $64$. 
This suggests that PriFT preserves a broader set of successful reasoning trajectories, achieving strong performance across different sampling budgets.

\subsection{\methodname reduces distribution shift compared with online reweighting}

Prior work shows that DFT can induce a large distribution shift from the pretrained model during fine-tuning \citep{zhu2025asft}. 
We show that \methodname reduces this shift without explicit KL regularization, by assigning less training signal to tokens weakly supported by the pretrained distribution. 
\autoref{fig:pred_entropy}(c) tracks the average token-level KL divergence from the pretrained distribution to the online distribution when fine-tuning a \texttt{LLaMA-2-7B model} on 10k data from MedMCQA \citep{pal2022medmcqa}. 
DFT shows a large and fast increasing KL divergence, as a result of its self-reinforcing online reweighting dynamics. 
In contrast, PriFT-prob and PriFT-mass keep the KL substantially lower, remaining close to SFT throughout training. 
This suggests that pretrained-reference weighting helps avoid the large distribution drift caused by online reweighting. \looseness=-1

\subsection{Ablation studies}
\label{sec:ablation}

We summarize the main ablation results here, with additional details provided in Appendix~\ref{app:ablation}.

\textbf{Replacing pretrained reference with an EMA model.}
As shown in \autoref{fig:ablation}(a), performance generally improves with larger EMA momentum when replacing the pretrained signal with EMA model, but still does not surpass the pretrained-reference reweighting. This suggests that PriFT benefits from preserving original pretrained distribution, rather than merely smoothing the online model. \looseness=-1

\textbf{Varying token selection threshold in PriFT-mass.}
We vary the selection threshold $\tau$ in PriFT-mass. As shown in \autoref{fig:ablation}(b), performance peaks around default threshold $\tau=0.5$. This indicates PriFT-mass benefits from filtering weakly supported tokens while retaining sufficient supervision. \looseness=-1

\textbf{Reweighting signal from a stronger model.}
We test if PriFT-prob benefits from reweighting signals provided by a stronger model. 
As shown in \autoref{tab:distill}, using a math-domain reference to reweight a base model yields mild performance gains, whereas a larger math-domain reference does not improve the smaller math-domain model. 
This suggests that a stronger reference does not necessarily provide a better reweighting signal. 
The model's own pretrained checkpoint can remain the most reliable source of prior support, as its distribution is best aligned with the model being fine-tuned. %

\textbf{Applying pretrained reweighting signals to existing baselines.}
We replace the online token-reweighting signals in EAFT \citep{diao2026eaft}, IDFT \citep{zhang2026idft}, and TALR \citep{lin2025talr} with their pretrained-reference counterparts while keeping the weighting rules unchanged. As shown in \autoref{tab:pretrained_weight_math_results}, this consistently improves average Pass@16, showing that pretrained-reference signals benefit not only for PriFT instantiations but also for other token-reweighing rules.

\textbf{Robustness across fine-tuning datasets.}
We further evaluate on two alternative fine-tuning datasets, \texttt{ScaleQuest-Math} \citep{ding2025unleashing} and \texttt{GSM8K} \citep{cobbe2021trainingmath}. 
PriFT consistently improves over SFT, although its advantage varies across training distributions: the best PriFT variant improves Avg@16 / Pass@16 by $6.62/1.52$ points on \texttt{ScaleQuest-Math}, $17.13/14.08$ points on the main \texttt{NuminaMath-CoT} setting, and $9.28/15.70$ points on \texttt{GSM8K}. 
These results show that PriFT is not tied to a particular fine-tuning dataset, while suggesting that the benefit of pretrained-reference weighting can depend on the underlying training distribution. Detailed results are provided in Appendix~\ref{app:dataset_robustness}. \looseness=-1

\textbf{Connection to knowledge distillation (KD).}
The distribution-matching term in KD decomposes into a target-token and a non-target term:
$
-\sum_v q_t(v)\log p_t(v)
=
-q_t(y_t)\log p_t(y_t)
-
\sum_{v\neq y_t} q_t(v)\log p_t(v),
$
where $q_t$ and $p_t$ denote the teacher and student distributions. 
While target term shares the form as PriFT-prob, we test whether the non-target term benefits PriFT-prob by training \texttt{Qwen2.5-Math-1.5B} with \texttt{Qwen2.5-Math-7B} as the teacher and varying the non-target term weight $\beta$ \footnote{We use a different teacher as using the same backbone as teacher would yield no distribution-matching signal at $\beta=1$.}. 
We note that this diagnostic excludes hard-label CE mixing and temperature scaling.
As shown in \autoref{tab:kd_decomp}, performance declines as $\beta$ increases, peaking at $\beta=0$, which corresponds to PriFT-prob with the teacher as reference.
This suggests that useful teacher guidance primarily comes from target-token reweighting rather than full distribution matching in this setting.\looseness=-1

\section{Conclusion and Limitations}
\label{sec:conclusion_limitations}

We presented PriFT, a token-reweighted SFT framework that derives token weights from a frozen pretrained reference instead of the online model being optimized. 
By exploiting the prior support from the pretrained model throughout fine-tuning, PriFT improves fine-tuning performance across mathematical reasoning, code generation, and medical question answering. 
It also provides a stronger initialization for subsequent RL, suggesting that pretrained-reference weighting can improve SFT while maintaining the sample diversity needed for further on-policy optimization. \looseness=-1

Our study has limitations and leaves open directions for future work. 
Our evaluation focuses on moderate-size open models, validating PriFT on larger models and broader applications remains future work. 
Furthermore, while PriFT demonstrates the benefits of pretrained prior support, extending its effectiveness on tasks with weaker prior support remains an interesting direction for future research.
\looseness=-1

\section*{Acknowledgements}
The authors thank Guillermo Ortiz-Jimenez, Nikolaos Dimitriadis, Alessandro Favero, Yulun Jiang, Skander Moalla, and the anonymous reviewers for constructive discussions and comments.

\bibliographystyle{plainnat}
\bibliography{references}

\newpage

\appendix

\section{Related Work}
\paragraph{Token-reweighted supervised fine-tuning.}
Recent work improves SFT by reweighting or selecting target tokens according to model-derived signals, rather than treating all tokens in a response equally. DFT and TALR use target-token probability to assign larger weights to high-confidence tokens~\citep{wu2025dft,lin2025talr}, while ProFit uses the same confidence signal for hard token selection~\citep{liu2026profit}. EAFT instead gates training with predictive entropy, emphasizing tokens whose predictions remain uncertain~\citep{diao2026eaft}. IDFT combines probability and entropy through a centered log-likelihood statistic to emphasize tokens with higher estimated learning signal~\citep{zhang2026idft}. Ranktuner takes into account the relative rank of each target label in the predictive distribution of the model \citep{yu2026ranktuner}.
\cite{li2025beyond} show that probability-based objectives favoring high-confidence tokens outperform standard SFT when pretrained models have strong task-relevant priors, whereas standard SFT remains preferable when such priors are weak.
ASFT further adds a pretrained-model anchoring term on top of probability-based reweighting, aiming to regularize the fine-tuned model toward the pretrained distribution~\citep{zhu2025asft}.
These methods differ in how they define token utility, but typically derive the reweighting signal from the online model being fine-tuned. Consequently, token weights evolve with the optimization trajectory and may reflect early preferences or overfitting. PriFT instead computes weights from a frozen pretrained reference, separating token-weight estimation from online adaptation and preserving a stable estimate of prior support. \looseness=-1

\paragraph{Knowledge distillation}
PriFT superficially resembles knowledge distillation (KD) \citep{hinton2015distilling}, in that token-level signals from a reference model influence the training of another model. However, standard KD typically transfers knowledge by matching a teacher distribution, whereas PriFT keeps the demonstration token as the supervised target and uses the reference model only to estimate how strongly that token is supported. 
A closer family is student-aware distillation, where the student's own behavior is used to shape the training signal. DistiLLM~\citep{ko2024distillm} and DistiLLM-2~\citep{ko2025distillm} exploit student-generated outputs for adaptive off-policy or contrastive distillation, while SODA~\citep{chen2026soda} constructs a contrastive signal from a one-time static snapshot of the base student's responses. These methods share with PriFT the view that the student's own distribution contains useful information about which supervision signals are relevant or learnable. PriFT differs by using this information at the token level to weight gold demonstrations, rather than using student outputs as distillation targets or contrastive negatives.
Finally, recent on-policy distillation methods address the train--test distribution mismatch of off-policy KD by training on student-generated trajectories with dense teacher feedback \citep{agarwal2024gkd,gu2024minillm,ko2024distillm,xu2025speculativekd,zhao2026opsd,shenfeld2026self}. PriFT shares their support-aware view, but takes the opposite route: instead of moving the training distribution toward student-generated states, it keeps offline supervision on high-quality demonstrations, and uses the initial checkpoint of the student to decide which demonstration tokens are supported by the student's own pre-adaptation prior. \looseness=-1

\paragraph{Comparison between SFT and RL}
Previous works have shown that supervised fine-tuning can distort pre-trained features, leading to catastrophic forgetting of pretrained knowledge \citep{kumar2022fine, wortsman2022robust, wang2025lines}.
A recent line of work argues that RL often generalizes better than SFT and exists less forgetting of the model's existing capabilities. \citet{chu2025sftmem} show that SFT tends to memorize surface patterns while RL transfers more broadly, and \citet{shenfeld2025rl} argue that on-policy RL forgets less precisely because it trains on rollouts the model itself produces. These findings motivate token-reweighting methods that try to make SFT behave more like on-policy training~\citep{qin2025iwsft,wu2025dft, zhang2026idft, zhu2025asft}.  \looseness=-1

\paragraph{SFT objectives as initialization for RL}
A complementary line of work argues that SFT objectives should be evaluated not only by the performance of the fine-tuned checkpoint, but also by the initialization they provide for subsequent RL. 
\citet{zhang2026pear} propose PEAR, which reweights offline SFT with importance ratios to reduce the offline--online distribution mismatch, and recent work on lightweight SFT similarly observes that overly aggressive supervised fitting can reduce output diversity and limit later RL improvement \citep{ligetting}. \cite{li2025preserving} propose GEM, a method to better preserve the sampling diversity during supervised fine-tuning stage, which can potentially improve exploration to improve performance limits with reinforcement learning. 
We adopt this perspective as an auxiliary evaluation of token-reweighted SFT. 
Our results show that PriFT improves over vanilla SFT and DFT both before RL and after applying the same RL procedure, suggesting that pretrained-reference-guided weighting strengthens the supervised checkpoint without exhausting the trajectory diversity and optimization headroom needed for later on-policy training.

\section{Baselines}

\paragraph{Baseline details.}
For consistency, we describe all token-reweighted baselines under a unified per-token objective. Let
$p_t=\pi_{\bm\theta_{\mathrm{on}}}(y_t\mid \bmx,y_{<t})$ denote the online model's probability of the target token. Standard weighted SFT minimizes
$$
\mathcal{L}(\theta_{\mathrm{on}})
=
\mathbb{E}_{(\bmx,\bmy)\sim\mathcal{D}}
\left[
-\sum_{t=1}^{T} m_t \log p_t
\right],
$$
where $m_t$ is the token-level weight. Different baselines instantiate $m_t$, or the corresponding per-token objective, as follows.

\begin{itemize}
    \item \textbf{SFT}.
    Standard supervised fine-tuning treats all target tokens equally with
    $m_t=1$. The per-token objective is therefore the standard negative log-likelihood,
    $\ell_t=-\log p_t$.

    \item \textbf{DFT} \citep{wu2025dft}.
    Dynamic Fine-Tuning rescales each token loss by the model's target-token probability:
    $m_t=\operatorname{sg}(p_t)$, where $\operatorname{sg}(\cdot)$ denotes stop-gradient. Its per-token objective is
    $\ell_t=-\operatorname{sg}(p_t)\log p_t$.
    This gives larger weights to tokens that the online model already predicts with higher confidence.

    \item \textbf{TALR} \citep{lin2025talr}.
    Token-Adaptive Loss Reweighting assigns adaptive weights according to token difficulty. Since $\ell_t=-\log p_t$, its unnormalized weight can be written as
    $m_t\propto \exp(-\ell_t/\tau)=p_t^{1/\tau}$,
    where $\tau$ is an adaptive temperature parameter. In practice, TALR applies stop-gradient to the weight and uses a lower cutoff $w_{\min}$:
    $m_t=\max\{\operatorname{sg}(p_t^{1/\tau}), w_{\min}\}$.
    This downweights low-probability tokens while preventing their weights from vanishing. In our experiments, we set $w_{\min}=0.01$.

    \item \textbf{ASFT} \citep{zhu2025asft}.
    Anchored Supervised Fine-Tuning extends DFT with a KL anchoring term to constrain the online model near a fixed reference model, typically the pretrained checkpoint. Its objective can be written as
    $$
    \ell_t
    =
    -\operatorname{sg}(p_t)\log p_t
    +
    \beta\,
    D_{\mathrm{KL}}\!\left(
    \pi_{\bm\theta_{\mathrm{pt}}}(\cdot\mid \bmx,y_{<t})
    \,\Vert\,
    \pi_{\bm\theta_{\mathrm{on}}}(\cdot\mid \bmx,y_{<t})
    \right),
    $$
    where $\beta$ controls the strength of the anchoring regularization, and we set $\beta=0.05$ in our experiments.

    \item \textbf{EAFT} \citep{diao2026eaft}.
    Entropy-Adaptive Fine-Tuning uses a normalized predictive-entropy score as a soft gating signal for token-level training. The per-token weight is calculated as $\widetilde{H}_t = H_t^{\mathrm{top}\text{-}K}/\log K$, where the token weights are detached and we set $K=20$ in our experiments. The resulting weight scales each token loss according to the model's uncertainty, suppressing low-entropy tokens while retaining supervision on uncertain tokens.

    \item \textbf{IDFT} \citep{zhang2026idft}.
      In-Distribution Fine-Tuning adaptively modifies probability-based token
      weights using an online score derived from target-token probability and
      predictive entropy. Compared with DFT, IDFT does not use raw target-token probability directly; instead, it adjusts the probability-based weight according to the token's signal to noise ratio. The per-token weight is calculated as $\text{sg}(p_t^{\exp(-\phi_t)})$ with $\phi_t=\log p_t+H_t$, where in our experiments
      $\phi_t$ is clipped to
      $[-b,b]$ with $b=1.0$.

\end{itemize}

\section{Full training and evaluation details}
\label{app:training_details}

\textbf{Mathematical reasoning. \ }
For mathematical reasoning, we follow the setup from \citet{wu2025dft}, and report results on various models.
Each model is fine-tuned for one epoch on the same 100k randomly sampled examples from \texttt{NuminaMath-CoT} \citep{li2024numinamath}.
We use the AdamW optimizer with a learning rate of $5\times10^{-5}$, cosine decay, and a warm-up ratio of 0.1. The batch size is 256, and the maximum sequence length is 2048 tokens.
We evaluate on \texttt{math\_oai} \citep{math500}, \texttt{minerva\_math} \citep{mineramath}, \texttt{olympiadbench} \citep{he2024olympiadbench}, \texttt{aime24}, and \texttt{amc23}, using the default chat template with chain-of-thought prompting.
Results are aggregated over 16 stochastic decoding runs with a temperature of 1.0 and a maximum generation length of 4096 tokens.
{Avg@16} denotes the mean accuracy across the 16 runs, and {Pass@16} denotes the fraction of examples solved by at least one sampled response.

\textbf{Code generation. \ }
For code generation, we fine-tune \texttt{Qwen2.5-Coder-3B} \citep{hui2024qwen25coder} and \texttt{Qwen3-4B-Instruct} \citep{yang2025qwen3} for one epoch on \texttt{Tulu-3-SFT-Personas-Code} \citep{lambert2024tulu3}, containing 34,999 examples focused on diverse Python coding questions.
We use the same optimizer, learning-rate schedule, batch size, and maximum sequence length as in mathematical reasoning.
We evaluate on four benchmarks: \texttt{HumanEval+} \citep{chen2021evaluatingcode,evalplus}, \texttt{MBPP+} \citep{austin2021mbpp,evalplus}, and two recent contest splits from \texttt{LiveCodeBench} \citep{jain2024livecodebench} (v5 and v6).
We report pass@1 accuracy using greedy decoding, with a maximum generation length of 768 tokens for \texttt{HumanEval+}/\texttt{MBPP+} and 2000 tokens for \texttt{LiveCodeBench}.

\section{Additional analyses}

\subsection{Additional analysis: comparison between DFT and standard SFT}
\label{app:sft_comparison}

\begin{figure}[t]
    \centering
    \includegraphics[width=1.0\linewidth]{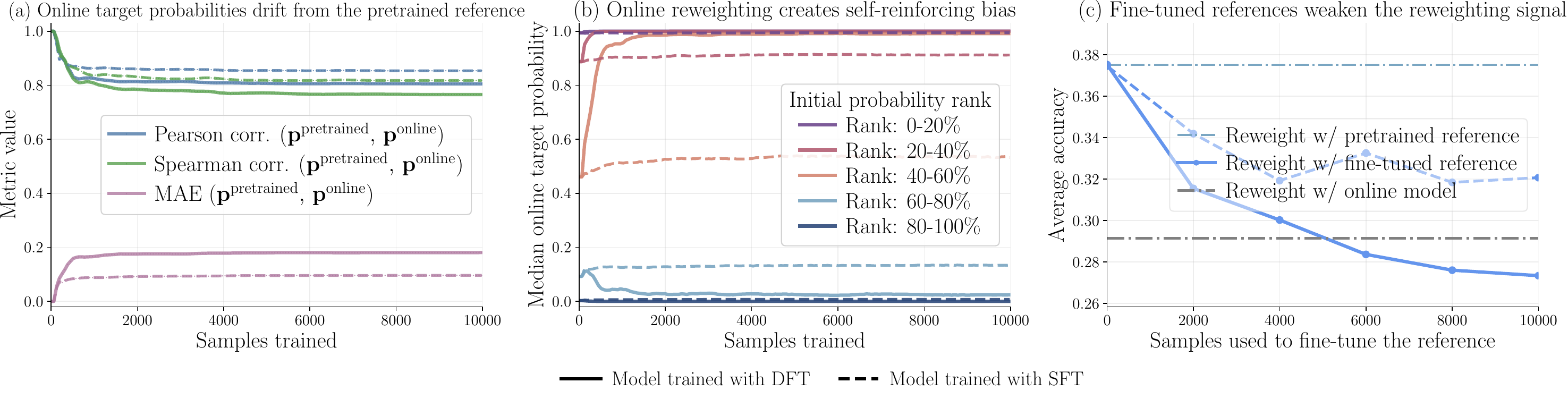}
    \caption{
    \textbf{Additional comparison between DFT and standard SFT for the analysis in \autoref{fig:pt_reference}.}
    The main text analyzes DFT as a representative online reweighting method; here we repeat the same measurements for a model trained with standard SFT.
    Solid curves correspond to DFT, and dashed curves correspond to SFT.
    (a) Both methods change the predictive distribution during fine-tuning, but SFT exhibits weaker drift from the pretrained reference.
    (b) The self-reinforcing rank bias is much stronger under DFT: high-support tokens are pushed more aggressively toward near-deterministic probabilities, while low-support tokens remain more suppressed.
    (c) Fine-tuned references degrade less severely under SFT than under DFT.
    This comparison shows that the instability observed in \autoref{fig:pt_reference} is amplified by online reweighting, rather than being solely caused by fine-tuning.
    }
    \label{fig:pt_reference_app}
\end{figure}

In Section~\ref{sec:analysis}, we analyze the instability induced by online reweighting using DFT as a representative example. 
Here, we repeat the same analysis for a model trained with standard SFT, which does not use online model predictions to construct token weights. 
As shown in \autoref{fig:pt_reference_app}, SFT also changes the predictive distribution during fine-tuning, but the drift from the pretrained reference is weaker than under DFT. 
The self-reinforcing rank bias is also less pronounced: high-support tokens are not pushed toward near-deterministic probabilities as aggressively, while low-support tokens remain less suppressed. 
In addition, when partially fine-tuned checkpoints are used as frozen references, the performance degradation is milder for SFT than for DFT.

These results suggest that the effects observed in Section~\ref{sec:analysis} are not merely caused by fine-tuning itself. 
They are amplified by online reweighting, where the model's evolving predictions directly determine future token weights. 
This further supports the use of a frozen pretrained model as a stable source of token-reweighting signals.

\subsection{Qualitative visualization of self-reinforcing token weights}
\label{app:self_reinforcing_visualization}

\begin{figure}[t]
    \centering
    \includegraphics[width=1.0\linewidth]{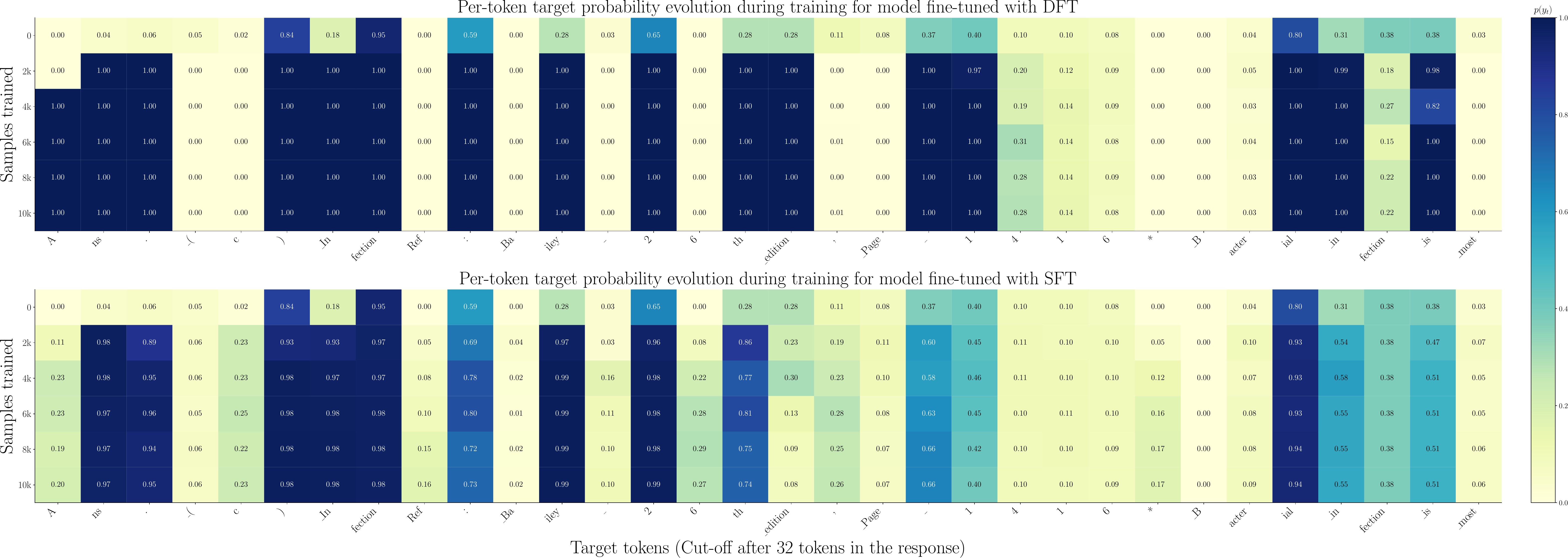}
    \caption{
    \textbf{Qualitative visualization of self-reinforcing token probabilities under DFT.}
    We track the target-token probability $p(y_t)$ of a representative training example across fine-tuning checkpoints.
    Columns correspond to target tokens, rows correspond to checkpoints, and colors indicate the online model's probability assigned to the target token.
    Compared with standard SFT, DFT rapidly pushes many initially high-probability tokens toward probability close to one, while low-probability tokens remain close to zero.
    This illustrates how online probability-based reweighting amplifies the model's early token preferences and produces a more polarized learning signal.
    }
    \label{fig:token_prob_heatmap}
\end{figure}

To further illustrate the self-reinforcing behavior of online reweighting, we visualize the target-token probabilities of a representative held-out example. 
\autoref{fig:token_prob_heatmap} compares a model trained with DFT and a model trained with standard SFT. 
Each column corresponds to a target token, and each row corresponds to a training checkpoint. 
Under DFT, many initially high-probability tokens are rapidly pushed toward probability close to one, while many low-probability tokens remain close to zero throughout training. 
This produces a polarized token-level pattern, where early preferences of the model are amplified into near-deterministic predictions. 

In contrast, standard SFT exhibits a less extreme evolution of target-token probabilities. 
Although SFT also changes the predictive distribution, it does not use the online probabilities themselves as token weights, and therefore shows weaker probability polarization. 
This qualitative example supports the analysis in Section~\ref{sec:pretrained_reference}: online probability-based reweighting can create a feedback loop in which tokens favored early in training receive larger updates. As a result of this rich-get-richer dynamics, tokens with lower initial weights do not get enough supervision during training.

\section{Additional results}
\label{sec:additional_results}

\subsection{Robustness to token-weight normalization}
\label{app:weight_normalization}

Following the original implementations of the token-reweighted baselines, our main experiments apply token weights directly without additional normalization. 
However, different weighting rules can induce different average loss scales and therefore different effective learning rates. 
To test whether this confounds the comparison, we repeat the \texttt{Qwen2.5-Math-7B} experiments while normalizing token weights to have mean $1$, either within each batch or within each sequence. 
All other training and evaluation settings are kept unchanged.

\begin{table}[t]
\centering
\small
\setlength{\tabcolsep}{3.5pt}
\caption{
Effect of token-weight normalization on \texttt{Qwen2.5-Math-7B}. 
Each entry reports Avg@16 / Pass@16 averaged over the five mathematical reasoning benchmarks.
}
\label{tab:weight_normalization}
\resizebox{0.6\linewidth}{!}{%
\begin{tabular}{lccc}
\toprule
\textbf{Method} & \textbf{Unnormalized} & \textbf{Per-batch} & \textbf{Per-sequence} \\
\midrule
SFT             & 23.54 / 55.52 & 23.54 / 55.52 & 23.54 / 55.52 \\
DFT             & 35.89 / 53.86 & 35.77 / 57.81 & 37.45 / 56.47 \\
IDFT            & 35.46 / 54.17 & 38.18 / 55.43 & 36.17 / 52.51 \\
TALR            & 36.70 / 58.06 & 37.22 / 59.46 & 35.61 / 59.04 \\
ASFT            & 35.90 / 60.85 & 35.89 / 63.55 & 35.68 / 64.33 \\
EAFT            & 23.06 / 56.75 & 23.33 / 55.41 & 23.69 / 58.13 \\
\rowcolor{blue!5}
\methodname-prob & 39.65 / 68.62 & 39.32 / \textbf{68.25} & 39.78 / \textbf{70.14} \\
\rowcolor{blue!5}
\methodname-mass & \textbf{40.67} / \textbf{69.60} & \textbf{40.16} / 66.36 & \textbf{40.54} / 67.45 \\
\bottomrule
\end{tabular}%
}
\end{table}

As shown in \autoref{tab:weight_normalization}, both PriFT variants remain the two strongest methods on both Avg@16 and Pass@16 under either normalization scheme. 
In particular, per-sequence normalization yields $39.78$ Avg@16 and $70.14$ Pass@16 for PriFT-prob, while PriFT-mass achieves the best Avg@16 of $40.54$. 
These results show that the improvements of PriFT cannot be explained by differences in the overall token-weight scale or effective learning rate. \looseness=-1

\subsection{Robustness across random seeds}
\label{app:seed_robustness}

To evaluate the robustness of the reported gains to randomness in training and evaluation, we repeat the \texttt{Qwen2.5-Math-7B} experiments using three independent random seeds and report the mean and standard deviation across runs in \autoref{tab:seed_robustness}.

\begin{table}[t]
\centering
\small
\setlength{\tabcolsep}{6pt}
\caption{
Three-seed results on \texttt{Qwen2.5-Math-7B}. 
We report mean $\pm$ standard deviation across independent runs.
}
\label{tab:seed_robustness}
\begin{tabular}{lcc}
\toprule
\textbf{Method} & \textbf{Avg@16} & \textbf{Pass@16} \\
\midrule
SFT             & $23.28 \pm 0.36$ & $55.29 \pm 0.92$ \\
DFT             & $36.31 \pm 0.40$ & $55.02 \pm 1.12$ \\
EAFT            & $22.78 \pm 0.61$ & $56.39 \pm 0.48$ \\
IDFT            & $35.84 \pm 1.02$ & $53.40 \pm 1.31$ \\
TALR            & $36.57 \pm 0.21$ & $58.73 \pm 0.66$ \\
ASFT            & $35.97 \pm 0.25$ & $62.23 \pm 1.21$ \\
\rowcolor{blue!5}
\methodname-prob & $39.22 \pm 0.42$ & $68.49 \pm 1.01$ \\
\rowcolor{blue!5}
\methodname-mass & $\mathbf{39.82} \pm 0.82$ & $\mathbf{68.72} \pm 0.93$ \\
\bottomrule
\end{tabular}
\end{table}

Both PriFT variants consistently outperform the baselines in the three-run averages. 
PriFT-mass improves over the strongest baseline mean by $3.25$ points on Avg@16 and $6.49$ points on Pass@16, while the observed run-to-run variation remains small relative to these margins. 
This confirms that the main conclusions are robust across independent random seeds. \looseness=-1

\subsection{Scaling to a 14B model}
\label{app:14b_scaling}

We further evaluate PriFT at a larger model scale using \texttt{Qwen2.5-14B}. 
We compare the pretrained checkpoint, standard SFT, DFT, PriFT-prob, and PriFT-mass on the same five mathematical reasoning benchmarks used in the main experiments.

\begin{table}[t]
\centering
\small
\setlength{\tabcolsep}{7pt}
\caption{
Mathematical reasoning results on \texttt{Qwen2.5-14B}, averaged over the five evaluation benchmarks.
}
\label{tab:qwen25_14b}
\begin{tabular}{lcc}
\toprule
\textbf{Method} & \textbf{Avg@16} & \textbf{Pass@16} \\
\midrule
Pretrained      & 21.47 & 57.96 \\
SFT             & 22.44 & 55.80 \\
DFT             & 32.54 & 54.32 \\
\rowcolor{blue!5}
\methodname-prob & 32.75 & \textbf{63.03} \\
\rowcolor{blue!5}
\methodname-mass & \textbf{33.80} & 61.48 \\
\bottomrule
\end{tabular}
\end{table}

As shown in \autoref{tab:qwen25_14b}, the advantage of pretrained-reference weighting persists at the 14B scale. 
PriFT-mass achieves the best Avg@16 of $33.80$, while PriFT-prob achieves the best Pass@16 of $63.03$. 
Compared with DFT, PriFT-prob improves Pass@16 by $8.71$ points, and PriFT-mass improves Avg@16 by $1.26$ points while also substantially improving Pass@16. 
These results extend the main findings beyond the 7B--8B models used in the primary experiments. \looseness=-1

\subsection{Robustness across fine-tuning datasets}
\label{app:dataset_robustness}

Our main mathematical reasoning experiments use \texttt{NuminaMath-CoT} for fine-tuning. 
To test whether the gains depend on this particular training distribution, we repeat the \texttt{Qwen2.5-Math-7B} experiments with two substantially different datasets: $100$k examples from \texttt{ScaleQuest-Math}  \citep{ding2025unleashing} and $7.4$k examples from \texttt{GSM8K} \citep{cobbe2021trainingmath}. 
\texttt{ScaleQuest-Math} is generated by \texttt{Qwen2-Math-7B-Instruct}, a model from the same family as our evaluated model, and may therefore exhibit a smaller distribution mismatch with the target model. 
In contrast, \texttt{GSM8K} is a much smaller and older dataset focused on grade-school mathematics, providing a substantially different fine-tuning distribution. 
We evaluate the resulting models on the same five mathematical reasoning benchmarks as in the main experiments.

\begin{table}[t]
\centering
\small
\setlength{\tabcolsep}{4.5pt}
\caption{
Results with alternative mathematical fine-tuning datasets on \texttt{Qwen2.5-Math-7B}. 
We report the average over the five evaluation benchmarks.
}
\label{tab:alternative_math_data}
\begin{tabular}{llcc}
\toprule
\textbf{Training data} & \textbf{Method} & \textbf{Avg@16} & \textbf{Pass@16} \\
\midrule
\multirow{4}{*}{ScaleQuest-Math}
& SFT              & 40.08 & 68.44 \\
& DFT              & 43.60 & 62.72 \\
& \cellcolor{blue!5}\methodname-prob 
& \cellcolor{blue!5}\textbf{46.70} 
& \cellcolor{blue!5}\textbf{69.96} \\
& \cellcolor{blue!5}\methodname-mass 
& \cellcolor{blue!5}45.82 
& \cellcolor{blue!5}68.44 \\
\midrule
\multirow{4}{*}{GSM8K}
& SFT              & 9.50 & 37.32 \\
& DFT              & 15.66 & 42.84 \\
& \cellcolor{blue!5}\methodname-prob 
& \cellcolor{blue!5}14.40 
& \cellcolor{blue!5}49.42 \\
& \cellcolor{blue!5}\methodname-mass 
& \cellcolor{blue!5}\textbf{18.78} 
& \cellcolor{blue!5}\textbf{53.02} \\
\bottomrule
\end{tabular}
\end{table}

As shown in \autoref{tab:alternative_math_data}, the choice of fine-tuning data substantially affects the absolute performance of all methods. 
Fine-tuning on \texttt{ScaleQuest-Math} produces a particularly strong SFT baseline, reaching $40.08$ Avg@16 and $68.44$ Pass@16. 
PriFT-prob further improves these results by $6.62$ Avg@16 and $1.52$ Pass@16 points, while PriFT-mass improves Avg@16 by $5.74$ points and maintains the same Pass@16. 
In comparison, the relative benefit of PriFT over SFT is substantially larger with the other training distributions. 
On the main \texttt{NuminaMath-CoT} setting, PriFT-mass improves over SFT by $17.13$ Avg@16 and $14.08$ Pass@16 points. 
When fine-tuning on \texttt{GSM8K}, PriFT-mass improves over SFT by $9.28$ Avg@16 and $15.70$ Pass@16 points, while PriFT-prob improves Pass@16 by $12.10$ points. \looseness=-1

This comparison is consistent with the role of prior support in controlling adaptation to offline demonstrations. 
When the fine-tuning data are already relatively compatible with the model, as the case for \texttt{ScaleQuest-Math}, standard SFT performs strongly and the additional benefit of pretrained-reference weighting is relatively smaller. 
For training distributions where SFT generalizes less effectively, PriFT provides substantially larger gains, particularly in Pass@16, suggesting that filtering or downweighting weakly supported supervision becomes more valuable. 
Nevertheless, the consistent improvements across all three training datasets show that PriFT is not tied to a particular fine-tuning distribution, while the varying gains are consistent with its motivation of reducing harmful adaptation to poorly supported offline targets. \looseness=-1

\subsection{Ablation study}
\label{app:ablation}

\begin{figure}[t]
    \centering
    \includegraphics[width=1.0\linewidth]{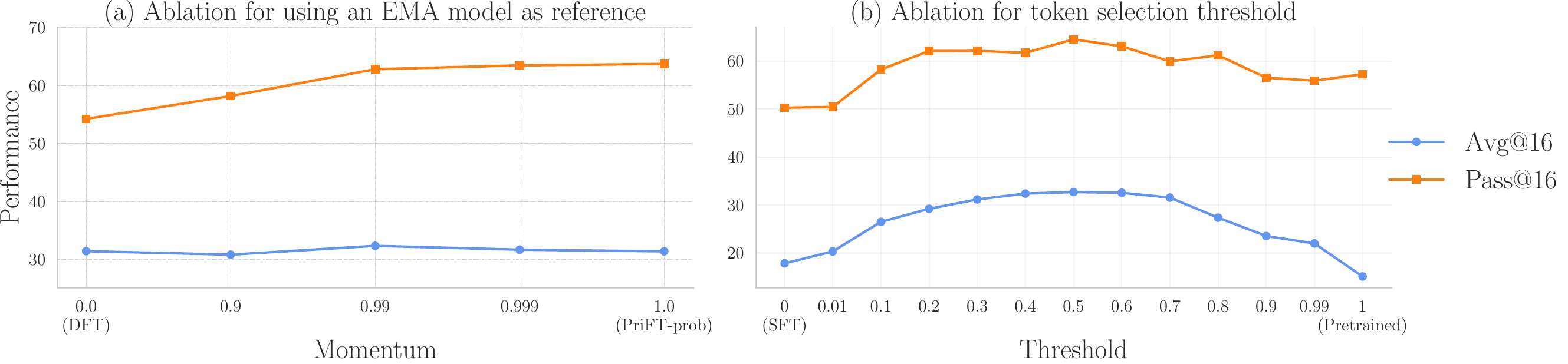}
    \caption{
    Ablation study on mathematical reasoning task with \texttt{Qwen2.5-Math-1.5B}.
    (a) Replace the pretrained reference in PriFT-prob with an EMA reference. 
    (b) Sensitivity of PriFT-mass to the selection threshold.
    }
    \label{fig:ablation}
\end{figure}

We conduct several ablations to better understand the design choices in PriFT. 
The results are shown in \autoref{fig:ablation} and \autoref{tab:distill}. Unless otherwise specified, experiments are conducted on \texttt{Qwen2.5-Math-1.5B}.

\paragraph{Replacing the pretrained reference with an EMA model.}
We test whether the frozen pretrained reference in PriFT-prob can be replaced by an exponential moving average (EMA) of the online model:
$
\bmth_{\mathrm{ema}}^{(k)}
\leftarrow
\rho \bmth_{\mathrm{ema}}^{(k-1)}
+
(1-\rho)\bmth_{\mathrm{on}}^{(k)}.
$
Here, $\rho=0$ recovers DFT, where token weights are computed from the current online model, while $\rho=1$ recovers PriFT-prob, where the reference remains the original pretrained model. 
As shown in \autoref{fig:ablation}(a), increasing the EMA momentum generally improves Pass@16, indicating that more stable token statistics are beneficial for weighted SFT. 
However, EMA references still do not outperform the pretrained-reference endpoint. 
This suggests that PriFT's advantage does not come only from reducing short-term fluctuations in the online model; it also depends on preserving the clean pretrained distribution before task-specific adaptation.

\paragraph{Varying the token selection threshold in PriFT-mass.}
We study the sensitivity of PriFT-mass to its selection threshold. 
Specifically, we replace the default threshold $0.5$ in \autoref{eq:ft_mass} with
$
m_t=\mathbf{1}\!\left[u_t^{(\textsc{mass})} > \tau\right].
$
When $\tau=0$, all tokens are selected and the objective reduces to standard SFT. 
When $\tau=1$, no tokens are selected, so the resulting model remains the pretrained checkpoint. 
\autoref{fig:ablation}(b) shows that performance improves from $\tau=0$, reaches its best value around $\tau=0.5$, and then declines as the selection becomes too strict. 
This supports the use of the default threshold: it filters out tokens with weak pretrained support while retaining enough supervised signal for effective adaptation. \looseness=-1

\paragraph{Reweighting signal from a stronger model.}
We test whether PriFT-prob benefit from leveraging the reweighting signal of a stronger model. 
In this setting, the online model is fine-tuned normally, but the reweighting signal comes from either its own pretrained checkpoint or another pretrained model. 
As shown in \autoref{tab:distill}, cross-model weighting can be beneficial when the reference provides more task-relevant prior knowledge. 
For \texttt{Qwen2.5-1.5B}, using the math-specialized Qwen2.5-Math-1.5B as the reweighting model improves both Avg@16 and Pass@16. 
However, this benefit is not uniform: for \texttt{Qwen2.5-Math-1.5B}, its own checkpoint outperforms using the larger \texttt{Qwen2.5-Math-7B} as the reference. 
These results suggest that external references can provide useful prior-support signals, but their effectiveness depends on both task relevance and compatibility with the model being fine-tuned.

\paragraph{Applying pretrained reweighting signals to existing baselines}

To test whether the benefit of pretrained statistics extends beyond PriFT, we modify existing token-reweighted SFT baselines by replacing their online weighting signal with statistics computed from the pretrained model. 
The weighting rule of each method is kept unchanged; only the source of the token statistics is changed. 
This gives pretrained-reweighting variants of EAFT \citep{diao2026eaft}, IDFT \citep{zhang2026idft}, and TALR \citep{lin2025talr}.
\begin{table*}[t]
\caption{Performance comparison between original and pretrained-weight variants on five mathematical reasoning benchmarks for \texttt{Qwen2.5-Math-1.5B} with 3 different baseline methods: EAFT, IDFT and TALR. We also report the average across all five benchmarks.}
\label{tab:pretrained_weight_math_results}
\vspace{0.4em}
\centering
\small
\setlength{\tabcolsep}{4.5pt}
\renewcommand{\arraystretch}{1.12}
\resizebox{\textwidth}{!}{%
\begin{tabular}{clcccccccccccc}
\toprule
\multirow{2}{*}{\textbf{Method}} & \multirow{2}{*}{\textbf{Variant}}
& \multicolumn{2}{c}{\textbf{MATH-OAI}}
& \multicolumn{2}{c}{\textbf{Minerva Math}}
& \multicolumn{2}{c}{\textbf{OlympiadBench}}
& \multicolumn{2}{c}{\textbf{AIME24}}
& \multicolumn{2}{c}{\textbf{AMC23}}
& \multicolumn{2}{c}{\textbf{Average}} \\
\cmidrule(lr){3-4}
\cmidrule(lr){5-6}
\cmidrule(lr){7-8}
\cmidrule(lr){9-10}
\cmidrule(lr){11-12}
\cmidrule(lr){13-14}
& & Avg@16 & P@16
  & Avg@16 & P@16
  & Avg@16 & P@16
  & Avg@16 & P@16
  & Avg@16 & P@16
  & Avg@16 & P@16 \\
\midrule

\multirow{2}{*}{EAFT}
& original           & 42.88 & 81.20 & \textbf{12.41} & \textbf{40.81} & \textbf{12.79} & \textbf{45.78} & 1.24 & 13.33 & 18.59 & \textbf{70.00} & 17.58 & 50.22 \\
& pretrained-weight & \textbf{43.53} & \textbf{82.00} & 11.40 & \textbf{40.81} & 12.36 & 45.63 & \textbf{2.08} & \textbf{16.67} & \textbf{19.38} & \textbf{70.00} & \textbf{17.75} & \textbf{51.02} \\
\midrule

\multirow{2}{*}{IDFT}
& original           & \textbf{66.11} & 84.80 & \textbf{22.21} & 42.65 & 27.54 & 52.59 & 7.31 & 20.00 & 31.72 & 72.50 & 30.98 & 54.51 \\
& pretrained-weight & 63.44 & \textbf{89.80} & 18.92 & \textbf{48.53} & \textbf{28.99} & \textbf{58.81} & \textbf{8.95} & \textbf{23.33} & \textbf{38.91} & \textbf{80.00} & \textbf{31.84} & \textbf{60.10} \\
\midrule

\multirow{2}{*}{TALR}
& original           & \textbf{64.08} & 89.00 & \textbf{22.96} & 44.85 & 27.73 & 55.70 & 6.26 & 20.00 & 38.75 & 77.50 & \textbf{31.95} & 57.41 \\
& pretrained-weight & 61.54 & \textbf{91.80} & 16.82 & \textbf{48.53} & \textbf{27.89} & \textbf{58.81} & \textbf{7.08} & \textbf{23.33} & \textbf{40.31} & \textbf{87.50} & 30.73 & \textbf{62.00} \\
\bottomrule
\end{tabular}%
}
\end{table*}

\autoref{tab:pretrained_weight_math_results} reports results on five mathematical reasoning benchmarks. 
Across all three methods, using pretrained statistics improves average Pass@16. 
These results show that replacing online statistics with pretrained statistics often improves the diversity and coverage of correct sampled trajectories, as reflected by Pass@16. 
Overall, this ablation supports the broader claim that the pretrained model provide a useful reweighting signal beyond the specific PriFT instantiations.

\paragraph{Connection to knowledge distillation (KD).}
We further analyze the relation between PriFT-prob and teacher-forced KD. 
Let $q_t$ and $p_t$ denote the teacher and student distributions at position $t$, respectively. 
The KD distribution-matching term can be decomposed into a target-token term and a non-target term:
$
-\sum_v q_t(v)\log p_t(v)
=
-q_t(y_t)\log p_t(y_t)
-
\sum_{v\neq y_t} q_t(v)\log p_t(v).
$
The target term has the same form as PriFT-prob: the reference model reweights the gold token by its probability, while the non-target term matches the teacher's probabilities on other tokens.

To test whether the non-target term provides useful guidance, we train \texttt{Qwen2.5-Math-1.5B} with \texttt{Qwen2.5-Math-7B} as the teacher and introduce a coefficient $\beta$:
$
\mathcal{L}_{\beta}
=
-q_t(y_t)\log p_t(y_t)
-
\beta\sum_{v\neq y_t} q_t(v)\log p_t(v).
$
Here, $\beta=0$ recovers PriFT-prob using the teacher model as reference, while $\beta=1$ recovers the full teacher-forced KD distribution-matching term. 
This is a diagnostic decomposition of the KD term, not the standard KD setting with an additional hard-label CE loss or temperature scaling for the teacher distribution.

\begin{table}[t]
\centering
\small
\caption{
Effect of the non-target term in decomposed teacher-forced KD. 
We use \texttt{Qwen2.5-Math-7B} as the teacher and \texttt{Qwen2.5-Math-1.5B} as the student. 
$\beta$ controls the strength of matching the teacher's non-target distribution.
}
\label{tab:kd_decomp}
\begin{tabular}{cccccc}
\toprule
 Non-target term weight $\beta$ & $0.00$ & $0.25$ & $0.50$ & $0.75$ & $1.00$ \\
\midrule
Avg@16  & \textbf{30.44} & 21.85 & 17.07 & 13.56 & 11.80 \\
Pass@16 & \textbf{59.93} & 58.74 & 54.92 & 52.17 & 48.64 \\
\bottomrule
\end{tabular}
\end{table}

As shown in \autoref{tab:kd_decomp}, performance decreases monotonically as $\beta$ increases, with the best result at $\beta=0$, which recovers PriFT-prob using the teacher model as reference. 
We note that result should not be interpreted as a comparison against standard KD. 
Rather, it isolates the KD distribution-matching term without an additional hard-label CE loss, and suggests that, in this controlled setting, useful teacher guidance mainly comes from target-label reweighting rather than matching the non-target distribution.

\begin{table}[t]
\caption{Effect of using cross-model token weighting signals in PriFT-prob.}
\label{tab:distill}
\vspace{0.4em}
\centering
\small
\setlength{\tabcolsep}{6pt}
\renewcommand{\arraystretch}{1.12}
\begin{tabular}{@{}clcc@{}}
\toprule
\textbf{Model} 
& \textbf{Token reweighting model} 
& \textbf{Avg@16} 
& \textbf{Pass@16} \\
\midrule

\multirow{2}{*}{\makecell[c]{Qwen2.5-Math\\1.5B}}
& Qwen2.5-Math-1.5B 
& \textbf{31.38} 
& \textbf{63.71} \\
& Qwen2.5-Math-7B 
& 30.44 
& 59.93 \\

\midrule

\multirow{2}{*}{\makecell[c]{Qwen2.5\\1.5B}}
& Qwen2.5-1.5B 
& 15.88 
& 46.13 \\
& Qwen2.5-Math-1.5B 
& \textbf{17.02} 
& \textbf{47.33} \\

\bottomrule
\end{tabular}
\end{table}

\subsection{Results on medical question answering}
\label{app:medical_results}

\paragraph{Experimental setup.}
For medical question answering, we follow the setup of \citet{zhu2025asft} in general. We fine-tune \texttt{LLaMA-2-7B} \citep{llama2} on 100k examples from \texttt{MedMCQA} \citep{pal2022medmcqa} for 1 epoch. We evaluate on \texttt{MedQA} \citep{medqa}, \texttt{MMLU-medical} \citep{mmlu}, and the \texttt{MedMCQA} test set. The model is trained for one epoch with a maximum sequence length of 512, a global batch size of 64, and a learning rate of $2 \times 10^{-5}$. Evaluation uses standard multiple-choice prompt templates, and we report accuracy.

\paragraph{Experimental results} \autoref{tab:medical_benchmark} reports results on medical question answering with \texttt{LLaMA-2-7B}. 
PriFT-mass achieves the best performance on MMLU and MedMCQA, with $49.14$ and $39.78$ accuracy, respectively. 
It also obtains the second-best average accuracy of $42.03$, closely matching ASFT, which achieves $42.40$ on average. 
Compared with other token-reweighted SFT baselines, PriFT-mass gives the strongest overall performance together with ASFT.

PriFT-prob also performs competitively, achieving the third-best average accuracy among all methods, over $6$ points higher than DFT. 
The gap between PriFT-prob and PriFT-mass is consistent with our motivation for PriFT-mass: raw pretrained probability can be overly restrictive in knowledge-intensive domains, where important target tokens may have low initial probability. 
Overall, these results show that PriFT generalizes beyond mathematical reasoning and provides an effective pretrained-reference weighting signal for medical question answering.

\begin{table}[t]
  \centering
  \setlength{\tabcolsep}{6pt}
  \caption{Performance on medical question answering benchmarks using \texttt{LLaMA-2-7B} fine-tuned on MedMCQA. We report accuracy on MedQA, MMLU-Medical, MedMCQA, and their average.}
  \resizebox{0.57\linewidth}{!}{%
  \begin{tabular}{lcccc}
  \toprule
  \textbf{Methods} & \textbf{MedQA} & \textbf{MMLU} & \textbf{MedMCQA} & \textbf{Avg.} \\
  \midrule
  Pretrained & 29.85 & 30.52 & 33.76 & 31.38    \\
  SFT        & 31.42 & 33.48 & 35.67 & 33.52 \\
  DFT        & 32.99 & 31.00 & 30.79 & 31.59 \\
  EAFT       & 32.29 & 30.34 & 31.17 & 31.27 \\
  IDFT       & 32.44 & 38.23 & 35.12 & 35.26 \\
  TALR       & 31.81 & 41.04 & 34.35 & 35.73 \\
  ASFT       & \textbf{40.93} & \underline{46.99} & \underline{39.28} & \textbf{42.40} \\
  \rowcolor{blue!5}
  PriFT-prob & 35.82 & 41.51 & 37.01 & 38.11 \\
  \rowcolor{blue!5}
  PriFT-mass & \underline{37.16} & \textbf{49.14} & \textbf{39.78} & \underline{42.03} \\
  \bottomrule
  \end{tabular}%
  }
  \label{tab:medical_benchmark}
\end{table}

\subsection{Results on mathematical reasoning: additional models}

\begin{table*}[t]
\caption{Performance comparison on five mathematical reasoning benchmarks. We also report the average across all five benchmarks.}
\centering
\small
\setlength{\tabcolsep}{4.5pt}
\renewcommand{\arraystretch}{1.12}
\resizebox{\textwidth}{!}{%
\begin{tabular}{llcccccccccccc}
\toprule
\multirow{2}{*}{\textbf{Model}} & \multirow{2}{*}{\textbf{Method}}
& \multicolumn{2}{c}{\textbf{MATH-OAI}}
& \multicolumn{2}{c}{\textbf{Minerva Math}}
& \multicolumn{2}{c}{\textbf{OlympiadBench}}
& \multicolumn{2}{c}{\textbf{AIME24}}
& \multicolumn{2}{c}{\textbf{AMC23}}
& \multicolumn{2}{c}{\textbf{Average}} \\
\cmidrule(lr){3-4}
\cmidrule(lr){5-6}
\cmidrule(lr){7-8}
\cmidrule(lr){9-10}
\cmidrule(lr){11-12}
\cmidrule(lr){13-14}
& & Avg@16 & P@16
  & Avg@16 & P@16
  & Avg@16 & P@16
  & Avg@16 & P@16
  & Avg@16 & P@16
  & Avg@16 & P@16 \\
\midrule

\multirow{8}{*}{\makecell[c]{Qwen2.5-Math\\(1.5B)}}
& Original & 30.19 & 84.20 & 8.56 & 39.71 & 15.82 & 53.19 & 4.99 & 26.67 & 15.94 & 82.50 & 15.10 & 57.25 \\
\cmidrule(lr){2-14}
& SFT      & 43.38 & 82.00 & 12.63 & 41.91 & 12.77 & 44.15 & 1.04 & 13.33 & 19.38 & 70.00 & 17.84 & 50.28 \\
& DFT      & \underline{64.21} & 87.20 & 21.74 & 44.12 & 27.35 & 53.19 & 5.63 & 16.67 & 38.13 & 70.00 & 31.41 & 54.23 \\
& EAFT     & 42.88 & 81.20 & 12.41 & 40.81 & 12.79 & 45.78 & 1.24 & 13.33 & 18.59 & 70.00 & 17.58 & 50.22 \\
& IDFT     & \textbf{66.11} & 84.80 & \underline{22.21} & 42.65 & 27.54 & 52.59 & \underline{7.31} & 20.00 & 31.72 & 72.50 & 30.98 & 54.51 \\
& TALR     & 64.08 & 89.00 & \textbf{22.96} & 44.85 & 27.73 & 55.70 & 6.26 & 20.00 & 38.75 & 77.50 & \underline{31.95} & 57.41 \\
& ASFT     & 58.89 & 88.00 & 16.99 & 46.69 & 25.42 & 58.07 & 4.79 & 23.33 & 32.97 & 80.00 & 27.81 & 59.22 \\
\rowcolor{blue!5}
& \methodname-prob & 63.06 & \textbf{91.40} & 18.17 & \textbf{48.90} & \underline{28.44} & \textbf{60.74} & \textbf{8.33} & \underline{30.00} & \underline{38.91} & \underline{87.50} & 31.38 & \underline{63.71} \\
\rowcolor{blue!5}
& \methodname-mass & 62.35 & \underline{90.60} & 19.09 & \underline{47.79} & \textbf{29.06} & \underline{58.37} & 6.25 & \textbf{33.33} & \textbf{46.72} & \textbf{92.50} & \textbf{32.69} & \textbf{64.52} \\
\midrule

\multirow{3}{*}{\makecell[c]{\quad Qwen2.5\\\quad (1.5B)}}
& Pretrained & 4.04 & 36.40 & 1.61 & 16.54 & 1.67 & 17.78 & 0.21 & 3.33 & 2.50 & 22.50 & 2.00 & 19.31 \\
& SFT & 25.59 & 68.60 & 4.15 & 24.26 & 5.88 & 32.44 & 0.42 & 6.67 & 7.81 & 47.50 & 8.77 & 35.90 \\
& DFT & \textbf{46.19} & 74.00 & \textbf{12.49} & \underline{33.46} & \textbf{13.89} & 38.37 & \underline{1.86} & \underline{13.33} & \underline{17.97} & \underline{62.50} & \textbf{18.48} & 44.33 \\
\rowcolor{blue!5}
& \methodname-prob & 41.54 & \underline{78.40} & 9.01 & \textbf{34.19} & 12.76 & \underline{42.22} & 1.24 & \underline{13.33} & 14.84 & \underline{62.50} & 15.88 & \underline{46.13} \\
\rowcolor{blue!5}
& \methodname-mass
& \underline{43.33} & \textbf{79.00} & \underline{11.04} & \underline{33.46} & \underline{13.49} & \textbf{43.85} & \textbf{3.33} & \textbf{16.67} & \textbf{18.59} & \textbf{65.00} & \underline{17.96} & \textbf{47.59} \\
\midrule

\multirow{3}{*}{\makecell[c]{Qwen2.5-Instruct\\(1.5B)}}
& Pretrained & 45.89 & \textbf{83.60} & 12.06 & 37.13 & 14.31 & 46.22 & 1.66 & \underline{13.33} & 18.59 & \underline{67.50} & 18.50 & 49.56 \\
& SFT & 25.73 & 69.20 & 4.61 & 25.74 & 6.22 & 33.33 & 0.41 & 6.67 & 8.44 & 62.50 & 9.08 & 39.49 \\
& DFT & 45.55 & 75.00 & 12.37 & 30.15 & 14.30 & 40.44 & 1.66 & \underline{13.33} & 20.94 & 60.00 & 18.96 & 43.78 \\
\rowcolor{blue!5}
& \methodname-prob & \underline{49.53} & \textbf{83.60} & \underline{14.59} & \underline{40.07} & \underline{16.82} & \textbf{47.26} & \underline{2.28} & \textbf{16.67} & \underline{25.47} & \textbf{75.00} & \underline{21.74} & \textbf{52.52} \\
\rowcolor{blue!5}
& \methodname-mass
& \textbf{52.78} & \underline{83.40} & \textbf{16.84} & \textbf{40.81} & \textbf{18.81} & \underline{46.67} & \textbf{2.49} & \textbf{16.67} & \textbf{27.81} & \underline{67.50} & \textbf{23.75} & \underline{51.01} \\
\midrule

\multirow{3}{*}{\makecell[c]{DeepSeekMath\\(7B)}}
& Pretrained & 37.78 & 76.80 & 17.94 & \underline{49.63} & 10.33 & 41.63 & 0.41 & 6.67 & 14.38 & \underline{62.50} & 16.17 & 47.45 \\
& SFT & 32.91 & 74.40 & 11.14 & 40.07 & 7.59 & 38.81 & 0.41 & 6.67 & 12.03 & 55.00 & 12.82 & 42.99 \\
& DFT & \textbf{46.65} & 76.60 & 17.40 & 37.50 & \textbf{16.07} & 41.63 & \textbf{2.28} & \textbf{13.33} & \underline{20.47} & 57.50 & \textbf{20.57} & 45.31 \\
\rowcolor{blue!5}
& \methodname-prob & \underline{44.06} & \textbf{78.80} & \underline{19.14} & 49.26 & \underline{14.46} & \textbf{46.96} & \underline{0.62} & 6.67 & 19.69 & \textbf{65.00} & 19.59 & \textbf{49.34} \\
\rowcolor{blue!5}
& \methodname-mass
& 43.34 & \underline{77.40} & \textbf{20.51} & \textbf{50.00} & 14.38 & \underline{43.70} & \underline{0.62} & \underline{10.00} & \textbf{21.72} & 60.00 & \underline{20.11} & \underline{48.22} \\
\bottomrule
\end{tabular}%
}
\label{tab:app_math_results}
\end{table*}

Beyond the main results in \autoref{tab:main_math_results}, we further evaluate PriFT on four additional mathematical reasoning backbones: \texttt{Qwen2.5-Math-1.5B}, \texttt{Qwen2.5-1.5B}, \texttt{Qwen2.5-Instruct-1.5B}, and \texttt{DeepSeekMath-7B} \citep{shao2024deepseekmath}. The results are shown in \autoref{tab:app_math_results}. Due to computational constraints, we include the most representative baseline from the main experiments rather than exhaustively evaluating all token-reweighted methods on every additional backbone.

Overall, PriFT continues to provide strong performance across model families and scales. On \texttt{Qwen2.5-Math-1.5B}, PriFT-mass achieves the best average performance on both Avg@16 and Pass@16, outperforming SFT and prior token-reweighted baselines. On \texttt{Qwen2.5-Instruct-1.5B} and \texttt{DeepSeekMath-7B}, PriFT also achieves the strongest aggregate Avg@16 or Pass@16 among fine-tuning methods, showing that pretrained-reference weighting is effective beyond the main 7B--8B settings. \looseness=-1

Taken together, these additional experiments support the robustness of PriFT across different pretrained, math-specialized, and instruction-tuned backbones.

\subsection{Additional results on code generation}
\label{app:code_generation}

\autoref{tab:code_qwen25_coder_app} reports additional code generation results on \texttt{Qwen2.5-Coder-3B}. PriFT-mass achieves the best average performance and is the only fine-tuning method that improves over the original checkpoint on average. The gains again concentrate on LiveCodeBench, while PriFT-prob is less effective on this specialized coding model, suggesting that relative support in PriFT-mass can be more robust than raw pretrained probability.

\begin{table}[t]
\caption{Additional code generation results on \texttt{Qwen2.5-Coder-3B}. We report pass@1 accuracy and the average across all four benchmarks.}
\centering
\setlength{\tabcolsep}{3.5pt}
\resizebox{0.6\columnwidth}{!}{%
\begin{tabular}{lccccc}
\toprule
\textbf{Method}
& \textbf{HumanEval+} & \textbf{MBPP+} & \textbf{LCB\,v5} & \textbf{LCB\,v6} & \textbf{Avg} \\
\midrule
Original
& \textbf{78.66} & 63.49 & \underline{21.25} & \underline{20.47} & \underline{45.97} \\
SFT
& 70.12 & \textbf{66.67} & 17.05 & 16.02 & 42.46 \\
DFT
& 70.12 & 65.34 & 12.73 & 12.13 & 40.08 \\
IDFT
& 71.34 & \underline{65.87} & 16.36 & 15.45 & 42.26 \\
EAFT
& 73.17 & 64.81 & 17.61 & 16.59 & 43.05 \\
TALR
& 70.12 & 65.34 & 12.84 & 11.85 & 40.04 \\
ASFT
& 73.78 & \underline{66.40} & 20.80 & 20.09 & 45.27 \\
\rowcolor{blue!5}
\methodname-prob
& 75.00 & 62.43 & 17.73 & 16.78 & 42.98 \\
\rowcolor{blue!5}
\methodname-mass
& \underline{77.44} & 64.29 & \textbf{22.27} & \textbf{21.14} & \textbf{46.28} \\
\bottomrule
\end{tabular}
}
\label{tab:code_qwen25_coder_app}
\end{table}

\section{Additional information for RL experiments}
\label{app:rl_setup}
For mathematical reasoning, we perform DAPO reinforcement learning following \citep{yu2025dapo} with the \texttt{verl} recipe and GRPO advantage estimation \citep{shao2024deepseekmath}. We train on 17,398 deduplicated prompts from \texttt{dapo-math-17k-dedup}, using rule-based rewards that assign $+1$ to correct boxed answers and $-1$ to incorrect ones, with an additional penalty for overlong responses. Experiments are conducted on \texttt{Qwen2.5-Math-1.5B} and \texttt{Qwen3-8B-Base} using 4 H200 GPUs for 100 RL steps, with prompt and generation batch sizes of 128, a PPO mini-batch size of 32, one PPO epoch, learning rate $10^{-6}$, 10 warmup steps, weight decay 0.1, gradient clipping 1.0, temperature 1.0, top-$p$ 1.0, and an actor KL loss coefficient of 0.01. During RL rollouts, we sample 32 responses per prompt for \texttt{Qwen2.5-Math-1.5B} and 16 for \texttt{Qwen3-8B-Base}.

\section{Compute resources}
\label{app:compute_resources}

All experiments are conducted on NVIDIA H100 or H200 GPUs. Each training run uses 4 GPUs with data-parallel training, including the mathematical reasoning, code generation, medical question answering, ablation, and RL-initialization experiments.

For mathematical reasoning SFT, a \texttt{Qwen2.5-Math-1.5B} run takes approximately 1 hour on 4 H200 GPUs, corresponding to about 4 GPU-hours. A \texttt{Qwen2.5-Math-7B} run takes approximately 2 hours on 4 H200 GPUs, corresponding to about 8 GPU-hours. These estimates include the additional reference forward pass used by PriFT to precompute token weights. Standard 16-sample mathematical reasoning evaluation takes around 20 minutes when parallelized over 4 GPUs.

For RL-initialization experiments, each DAPO run takes approximately 7 hours on 4 GPUs. Due to limited computational budget, we used the same fixed computational resources for all RL experiments rather than training each model to full convergence.

\section{Social Impacts}

This paper presents work whose goal is to advance the field
of Machine Learning. There are many potential societal
consequences of our work, none of which we feel must be
specifically highlighted here.

\clearpage
\newpage

\end{document}